\pdfoutput=1

\documentclass[11pt]{article}

\usepackage{acl}

\usepackage{times}
\usepackage{latexsym}
\usepackage{graphicx}
\usepackage{colortbl}
\usepackage{multirow}
\usepackage{booktabs}
\usepackage{subcaption}
\usepackage[export]{adjustbox}
\usepackage{geometry}
\usepackage{array}
\usepackage{float}
\usepackage{placeins}
\usepackage{afterpage}

\usepackage{algorithm}
\usepackage{algpseudocode}
\usepackage{float}
\usepackage{pdflscape}
\usepackage{hyperref}
\usepackage[]{acronym}
\usepackage{amsmath}

\usepackage{amssymb}

\usepackage{comment}
\usepackage[T1]{fontenc}

\usepackage[utf8]{inputenc}

\usepackage{microtype}

\usepackage{inconsolata}

\usepackage{graphicx}

\title{Decoding Decoded:\\ Understanding Hyperparameter Effects in Open-Ended Text Generation}

\author{
  \textbf{Esteban Garces Arias\textsuperscript{1,2}},
  \textbf{Meimingwei Li\textsuperscript{1}},
  \textbf{Christian Heumann \textsuperscript{1}},
  \textbf{Matthias Aßenmacher \textsuperscript{1,2}}
\\
\\
  \textsuperscript{1}Department of Statistics, LMU Munich,
  \textsuperscript{2}Munich Center for Machine Learning (MCML)
\\
\\
  \small{
    \textbf{Correspondence:} \href{mailto:Esteban.GarcesArias@stat.uni-muenchen.de}{Esteban.GarcesArias@stat.uni-muenchen.de}
  }
}

\begin{document}

\maketitle

\begin{abstract} Decoding strategies for generative large language models (LLMs) are a critical but often underexplored aspect of text generation tasks. Guided by specific hyperparameters, these strategies aim to transform the raw probability distributions produced by language models into coherent, fluent text. In this study, we undertake a large-scale empirical assessment of a range of decoding methods, open-source LLMs, textual domains, and evaluation protocols to determine how hyperparameter choices shape the outputs. Our experiments include both factual (e.g., news) and creative (e.g., fiction) domains, and incorporate a broad suite of automatic evaluation metrics alongside human judgments. Through extensive sensitivity analyses, we distill practical recommendations for selecting and tuning hyperparameters, noting that optimal configurations vary across models and tasks. By synthesizing these insights, this study provides actionable guidance for refining decoding strategies, enabling researchers and practitioners to achieve higher-quality, more reliable, and context-appropriate text generation outcomes.
\end{abstract}

\begin{figure}[ht]
\centering
\includegraphics[width = 50mm, keepaspectratio]
{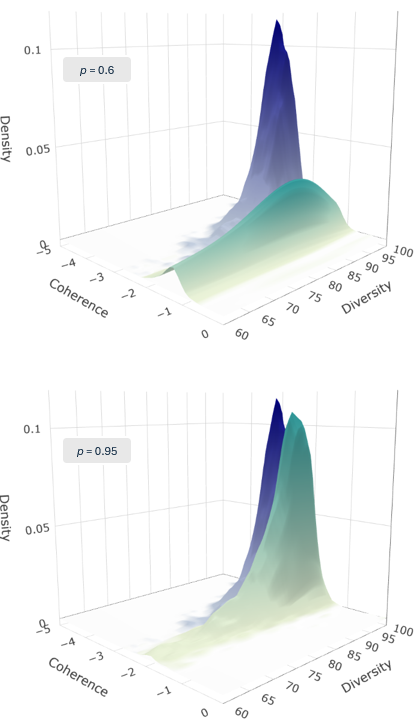}
\caption{Influence of the nucleus sampling hyperparameter $p$ on the distribution of diversity and coherence metrics in text generated by Mistral 7B v0.3 (green). For comparison, the distribution of the same metrics in human-written text is displayed in blue.}
\label{fig:figure_one}
\end{figure}

\section{Introduction}

Generative large language models (LLMs) do not directly produce natural language text. Instead, they generate a high-dimensional probability distribution over all tokens in their vocabulary. The process of converting these probabilities into coherent text, known as \textit{decoding}, can substantially influence the quality of the generated output, sometimes matching the impact of the LLM itself. 

Most decoding strategies employed with contemporary LLMs involve hyperparameters that play critical roles in shaping the generated text. These hyperparameters strongly influence factors such as coherence, fluency, and diversity \citep{zhou2024balancingdiversityriskllm}. Despite their importance, the selection and tuning of these hyperparameters remain under-explored. Users often rely on default settings and prioritize benchmarking different models over optimizing decoding strategies. This approach overlooks the varying requirements of different text generation tasks, ranging from factual domains like news generation to creative areas such as fiction. Additionally, the effectiveness of decoding strategies may vary across different models, a nuance that current practices— which assess strategies uniformly across models—fail to capture.

Recent research underscores the impact of hyperparameter configurations on both the coherence and diversity of generated text when using sampling-based decoding \citep{zhou2024balancingdiversityriskllm}. In this study, we extend this by systematically varying the hyperparameters of commonly used decoding strategies—deterministic, sampling-based, and contrastive—to evaluate their effects on text quality across diverse datasets, metrics, and open-source LLMs. By employing predefined grids for hyperparameters in deterministic (beam width), stochastic (top-$k$, top-$p$, and temperature), and contrastive ($\alpha$ and $k$) decoding strategies, we investigate their role in balancing coherence, fluency, diversity, and overall text quality. This investigation is particularly relevant for addressing common issues such as degeneration and ensuring effective adaptation to a wide range of text generation tasks \cite{shi2024thoroughexaminationdecodingmethods}. 

Furthermore, while modern LLMs are highly advanced, they remain susceptible to problems like incoherence or degeneration under certain hyperparameter settings. Our work addresses these challenges by providing a sensitivity analysis and offering practical guidelines for tuning decoding hyperparameters. This aims to optimize the desired properties of the generated text while mitigating undesirable behaviors, including repetitiveness, and inconsistencies or hallucinations.

\paragraph{Contributions} This study advances the field of decoding strategies for LLMs by performing a large-scale sensitivity analysis of commonly used decoding methods and examining their practical effects on model performance.

\begin{enumerate} 
    \item We conduct an extensive experimental analysis to evaluate the impact of hyperparameter values on various text quality metrics across models in open-ended text generation. 
    \item Our comprehensive analysis reveals key factors that influence the quality of LLM-generated texts, as assessed by widely adopted evaluation metrics covering multiple lexical dimensions. 
    \item Based on these insights, we offer actionable recommendations for practitioners to select appropriate decoding strategies and hyperparameters tailored to specific use cases. 
    \item We create (and share) a unique database for future research: In total, we generate \textbf{2.2 million} text continuations, which are publicly available for future meta-analyses, along with our complete codebase: \url{https://github.com/YecanLee/Decoding-Decoded}.
\end{enumerate}


\section{Decoding Strategies and Hyperparameters}
\label{sec:dec_strats}

Decoding strategies are generally categorized into two types: deterministic and stochastic. Given the complexity introduced by their hyperparameters, we additionally separate contrastive strategies (cf. Sec. \ref{sec:contr}) from the other deterministic ones (cf. Sec. \ref{sec:det}). While the latter are solely focused on maximizing the joint probability, the former explicitly compromise this objective by incentivizing more diverse texts. We further do not go into detail on other existing strategies that we do not employ, such as Frustratingly Simple Decoding \citep{yang-etal-2024-frustratingly}, Typical sampling \citep{meister2023locally}, Contrastive Decoding \citep{li2023contrastive}, and Adaptive  Decoding \citep{zhu2024improvingopenendedtextgeneration}.


\subsection{Deterministic Strategies}
\label{sec:det}

Deterministic strategies follow fixed decision-making processes that do not incorporate randomness. These methods are widely used in tasks requiring high reproducibility and reliability.

\paragraph{Greedy search.} This strategy selects the token with the highest probability at each time step, resulting in a sequence that maximizes the local likelihood. However, it often leads to suboptimal results due to the lack of look-ahead, which can cause the model to get trapped in less coherent output.

\paragraph{Beam search.} \citet{Freitag_2017} extends the greedy approach by maintaining a beam of the $w$ most probable sequences at each time step. The hyperparameter $w$, known as the beam width, aims to reduce the risk of suboptimal sequence choices, but the results can still generate repetitive text in open-ended tasks.


\subsection{Sampling-based Strategies}
\label{sec:sampl}

Sampling-based strategies introduce stochasticity to encourage diversity in text generation. These methods allow for more creative and varied outputs but require careful tuning of hyperparameters to balance coherence and randomness.

\paragraph{Sampling with temperature.} Introduced by \citet{ackley1985learning}, this method samples from the full distribution over all tokens while modifying the softmax. The temperature hyperparameter controls the sharpness of the distribution — higher temperatures flatten it, increasing randomness, while lower temperatures make it more deterministic, favoring tokens with higher probabilities even more.

\paragraph{Top-$k$ Sampling.} Proposed by \citet{fan2018hierarchical}, this strategy limits sampling to the top \( k \) tokens with the highest probabilities, forming a subset \( V^{(k)} \). Truncating the distribution reduces the risk of sampling from the long tail of low-probability (and potentially incoherent) tokens, while still introducing some diversity through stochasticity.

\paragraph{Nucleus Sampling.} Also known as top-$p$ sampling, this method was introduced by \citet{holtzman2019curious}. It samples from the smallest subset \( S \) whose cumulative probability exceeds a threshold \( p \). Nucleus sampling adapts dynamically to the token distribution, preserving both diversity and relevance by considering only the most probable tokens.


\subsection{Contrastive Strategies}
\label{sec:contr}

Contrastive strategies leverage comparisons between different models or hypotheses to improve text quality. These methods focus on enhancing coherence and informativeness without sacrificing diversity.

\paragraph{Contrastive Search.} \citet{su2022contrastive} introduce a look-ahead mechanism that penalizes tokens disrupting the isotropy of the latent space in the language model, penalizing degeneration while producing more semantically consistent text.

\paragraph{Adaptive Contrastive Search.} \citet{arias2024adaptivecontrastivesearchuncertaintyguided} propose an adaptive strategy that aims to strike a balance between coherence and diversity dynamically, based on the model entropy at each time step, removing the need for extensive hyperparameter tuning.


\section{Related Work}
\label{sec:related_work}

The selection of decoding strategies is critical in determining the performance of text generation models, particularly in balancing output quality and diversity. \citet{10.1162/tacl_a_00502} provide a foundational analysis of decoding methods, including beam search, top-$k$ sampling, and nucleus sampling. Their findings indicate that while beam search is effective for structured tasks such as machine translation, it often results in repetitive or less coherent text in creative tasks like story generation. This underscores the necessity of adapting decoding strategies to specific task requirements.

Subsequent studies, such as \citet{amini-etal-2023-generating}, build on these insights by offering a comprehensive review of the principles that guide text generation. This research highlights the influence of various decoding mechanisms on the final output, emphasizing the importance of selecting suitable strategies tailored to specific tasks. Additionally, it examines the trade-off between fluency and diversity, particularly in open-ended text generation scenarios. More recently, \citet{shi2024thoroughexaminationdecodingmethods} investigate how decoding strategies scale with large language models (LLMs), exploring the interactions between decoding strategies, their hyperparameters, and model size. The authors demonstrate the increasing complexity of optimizing these strategies as models grow in scale, especially in open-ended generation contexts. Our experiments substantially extend the work of \citet{shi2024thoroughexaminationdecodingmethods} by exploring additional hyperparameter combinations across a broader range of models and evaluating text quality using multiple metrics beyond MAUVE.

Additionally, \citet{zhou2024balancingdiversityriskllm} provide an in-depth exploration of sampling-based methods. They offer guidelines for managing the balance between diversity and the risk of incoherence, illustrating how hyperparameter tuning can influence the quality and diversity of generated outputs. This work is a practical resource for selecting decoding strategies based on task-specific requirements. In contrast, our focus extends beyond the evaluation of risk associated with tuning temperature and truncation parameters in stochastic strategies.

While providing valuable insights, many existing studies do not offer a comprehensive analysis of how different decoding strategies affect various quality metrics, such as coherence, diversity, and MAUVE. We address this gap and introduce \textit{QText}, which combines these metrics using the harmonic mean. We compare the best and worst-performing strategies based on these metrics with those identified by human evaluators, highlighting areas of agreement and divergence. Detailed insights are provided in Figure \ref{tab:human_eval} and Section \ref{appendix:performance_analysis}.


\section{Experimental setup}
\label{sec:experimental_setup}

We employ seven models to generate stories from prompts sourced from three distinct datasets, using six decoding strategies with varying hyperparameter configurations. To evaluate the quality of the generated text, we rely on three widely used automatic metrics: coherence, diversity, and MAUVE, and we also measure QText. In addition, human evaluators are engaged to provide further assessment of the quality of the generations.


\subsection{Models}

We employ GPT2-XL (1.5B parameters) \citep{radford2019language}, Mistral 7B v0.1 and v0.3 
\citep{touvron2023llama2openfoundation}, Llama 3.1 8B \citep{dubey2024llama3herdmodels}, Deepseek 7B \citep{deepseekai2024deepseekllmscalingopensource}, Qwen2 7B \citep{yang2024qwen2technicalreport}, and Falcon 2 11B \citep{malartic2024falcon211btechnicalreport}.

\begin{table*}[!ht]
\centering
\resizebox{1\textwidth}{!}{
\begin{tabular}{llllllr}
\hline
\textbf{Models} & \textbf{Datasets} & \textbf{Metrics} & \textbf{Decoding strategy}  & \textbf{Hyperparameter} & \textbf{Values}                  & \multicolumn{1}{l}{\textbf{\# Experiments}} \\ \hline
Deepseek        & Wikitext          & Coherence        & Beam search                 & Beam width              & \{3, 5, 10, 15, 20, 50\}      & 7 x 3 x 6 = 126                             \\
Falcon2          & Wikinews          & Diversity        & Contrastive search          & $k$                       & \{1, 3, 5, 10, 15, 20, 50\}      & 7 x 3 x 7 x 5 = 735                        \\
GPT2-XL         & Book              & MAUVE            &                             & $\alpha$                   & \{0.2, 0.4, 0.6, 0.8, 1.0\}      &                                             \\
Llama3          &                   & QText & Adaptive contrastive search & NA                      & NA                               & 1 x 3 x 1 = 3                              \\
Mistralv01         &                   &  Human Evaluation                & Sampling with temperature   & Temperature             & \{0.1, 0.3, 0.5, 0.7, 0.9, 1.0\} & 7 x 3 x 6 = 126                             \\
Mistralv03      &                   &                  & Top $k$ sampling              & $k$                       & \{1, 3, 5, 10, 15, 20, 50\}      & 7 x 3 x 7 = 147                             \\
Qwen2        &                   &                  & Top $p$ (nucleus) sampling    & $p$                       & \{0.6, 0.7, 0.8, 0.9, 0.95\}     & 7 x 3 x 5 = 105          \\ \hline
                &                   &                  &                             &                         & Grand Total                      & \multicolumn{1}{r}{1,242}                   
\end{tabular}
}
\caption{Overview of the experimental setup. The total number of experiments was determined by the combinations of models, datasets, and hyperparameter values explored. Falcon2 (11B) was run on an NVIDIA A100 GPU, while all other models were evaluated using an NVIDIA GeForce RTX 4090.}
\label{tab:experimental_setup}
\end{table*}

\subsection{Hyperparameters}
For contrastive search, we evaluate combinations of $\alpha \in \{0.2, 0.4, 0.6, 0.8, 1.0\}$ and $k \in \{1, 3, 5, 10, 15, 20, 50\}$, while for beam search, we consider beam width $\in \{3, 5, 10, 15, 20, 50\}$. For sampling with temperature we consider temperature $\in \{0.1, 0.3, 0.5, 0.7, 0.9, 1.0\}$, for top-$k$ sampling, we use $k \in \{1, 3, 5, 10, 15, 20, 50\}$ and for top-$p$ (nucleus) sampling we evaluate $p \in \{0.6, 0.7, 0.8, 0.9, 0.95\}$, for a total of 60 hyperparameter combinations. We also use the hyperparameter-free adaptive contrastive search for comparison, which increase the number of experiments by 3, totaling to 1242 experiments and 2.2 Mio generated stories. Further details are presented in Table \ref{tab:experimental_setup}.


\subsection{Evaluation Metrics}
\label{sec:metrics}

We follow \citet{su2022empirical} and use three metrics to automatically measure the quality the generations: Diversity, MAUVE, and Coherence.
\paragraph{Diversity.} This metric aggregates $\mathrm{n}$-gram repetition rates: $$\text{DIV}=\prod_{n=2}^4 \frac{\mid \text { unique } \mathrm{n} \text {-grams }\left(\mathrm{x}_{\text {cont }}\right) \mid}{\text { total } \mathrm{n} \text {-grams }\left(\mathrm{x}_{\text {cont }}\right) \mid}$$ A low diversity score suggests the model suffers from repetition, and a high diversity score means the model-generated text is lexically diverse.

\paragraph{MAUVE.} MAUVE \citep{pillutla2021mauve} score measures the distribution similarity between the set of generated text and the set of gold references.

\paragraph{Coherence.} Proposed by \citet{su2022contrastive}, the coherence metric is defined as the averaged log-likelihood of the generated text conditioned on the prefix text as

$$
\operatorname{Coherence}(\hat{\boldsymbol{x}}, \boldsymbol{x})=\frac{1}{|\hat{\boldsymbol{x}}|} \sum_{i=1}^{|\hat{\boldsymbol{x}}|} \log p_{\mathcal{M}}\left(\hat{\boldsymbol{x}}_i \mid\left[\boldsymbol{x}: \hat{\boldsymbol{x}}_{<i}\right]\right)
$$

where $\boldsymbol{x}$ and $\hat{\boldsymbol{x}}$ are the prefix text and the generated text, respectively; [:] is the concatenation operation and $\mathcal{M}$ is the OPT model (2.7B) \cite{zhang2022optopenpretrainedtransformer}. Finally, we apply a smoothed Min-Max normalization to the coherence values to ensure consistency and comparability with Diversity and MAUVE metrics.

\[
\operatorname{COH} = \frac{\text{Coherence} - \min(\text{Coherence}) + 1}{\max(\text{Coherence}) - \min(\text{Coherence}) + 1}
\]

\paragraph{Aggregation.} QText

Following a generalization of the F1-score for three metrics, we use the harmonic mean of Diversity, MAUVE, and Coherence as aggregation measure:

$$
\text{QText} = \frac{3}{\frac{1}{\text{DIV}} + \frac{1}{\text{MAUVE}} + \frac{1}{\text{COH}}}
$$

Values close to one indicate high-quality text generation, while values approaching zero reflect low-quality outcomes. 

\paragraph{Human Evaluation.} To evaluate the quality of the generated text, we consider two critical aspects: fluency and coherence. A fluent text is written in grammatical English and has a natural flow (e.g. excluding unnatural repetition or web formatting). A coherent text should stay on topic with the prompt and avoid unnatural topic drift. We provide five native English speakers with 39 competing continuations for one prompt per dataset, and subsequently ask them to rank them based on their perceived quality, for a total of 570 evaluations. Definitions and instructions for the rating process are shown in Appendix \ref{a:humeval}, Figure \ref{fig:human_evaluation_form}.


\subsection{Datasets}

We evaluate our proposed method across three domains in open-ended text generation: news, Wikipedia, and stories. For the news domain, we utilize 2,000 articles from Wikinews; for the Wikipedia domain, we employ 1,314 examples from the WikiText-103 dataset \citep[][]{merity2016pointer}; and for the stories domain, we employ the Project Gutenberg split of BookCorpus \citep[1,947 examples;][]{zhu2015aligning}. Each example consists of a prompt paired with a gold-standard, human-generated continuation for evaluation purposes.
For each prompt, we generate 256 tokens as a continuation. The resulting text is assessed using both automatic metrics (outlined in Section \ref{sec:metrics}
) and human scores.


\section{Results}
\label{sec:res}

\begin{table*}[!ht]
    \centering
    \resizebox{0.8\textwidth}{!}{
    \begin{tabular}{ccccccc}
                                                     &                                                                                                                       &                                    & \multicolumn{4}{c}{Automatic Evaluation}                  \\ \hline
\multicolumn{1}{c|}{Type}                            & \multicolumn{1}{c|}{Strategy}                                                                                         & \multicolumn{1}{c|}{Configuration} & Coherence & Diversity & MAUVE & QText \\ \hline
\multicolumn{1}{c|}{\multirow{42}{*}{Deterministic}} & \multicolumn{1}{c|}{\multirow{6}{*}{\begin{tabular}[c]{@{}c@{}}Beam search\\      (Beam width)\end{tabular}}}         & \multicolumn{1}{c|}{3}         & -0.72          & \textbf{12.78}          & \textbf{25.27}      & \textbf{21.50}      \\
\multicolumn{1}{c|}{}                                & \multicolumn{1}{c|}{}                                                                                                 & \multicolumn{1}{c|}{5}         & -0.69          & 12.14          & 22.01      & 19.92      \\
\multicolumn{1}{c|}{}                                & \multicolumn{1}{c|}{}                                                                                                 & \multicolumn{1}{c|}{10}        & -0.64          & 10.52          & 18.33      & 17.24      \\
\multicolumn{1}{c|}{}                                & \multicolumn{1}{c|}{}                                                                                                 & \multicolumn{1}{c|}{15}        & -0.62          & 9.40           & 16.37      & 15.59      \\
\multicolumn{1}{c|}{}                                & \multicolumn{1}{c|}{}                                                                                                 & \multicolumn{1}{c|}{20}        & -0.60          & 8.52           & 14.85      & 14.26      \\
\multicolumn{1}{c|}{}                                & \multicolumn{1}{c|}{}                                                                                                 & \multicolumn{1}{c|}{50}        & \textbf{-0.55}$\star$          & 6.17           & 9.94       & 10.19      \\ \cline{2-7} 
\multicolumn{1}{c|}{}                                & \multicolumn{1}{c|}{\multirow{35}{*}{\begin{tabular}[c]{@{}c@{}}Contrastive search\\      ($\alpha$, $k$)\end{tabular}}}       & \multicolumn{1}{c|}{(0.2, 1)}  & \textbf{-0.75}          & 9.90           & 24.21      & 18.25      \\
\multicolumn{1}{c|}{}                                & \multicolumn{1}{c|}{}                                                                                                 & \multicolumn{1}{c|}{(0.2, 3)}  & -0.95          & 20.73          & 42.30      & 33.95      \\
\multicolumn{1}{c|}{}                                & \multicolumn{1}{c|}{}                                                                                                 & \multicolumn{1}{c|}{(0.2, 5)}  & -0.98          & 22.75          & 44.22      & 36.44      \\
\multicolumn{1}{c|}{}                                & \multicolumn{1}{c|}{}                                                                                                 & \multicolumn{1}{c|}{(0.2, 10)} & -1.03          & 24.70          & 47.12      & 39.13      \\
\multicolumn{1}{c|}{}                                & \multicolumn{1}{c|}{}                                                                                                 & \multicolumn{1}{c|}{(0.2, 15)} & -1.07          & 25.82          & 47.85      & 40.32      \\
\multicolumn{1}{c|}{}                                & \multicolumn{1}{c|}{}                                                                                                 & \multicolumn{1}{c|}{(0.2, 20)} & -1.12          & 26.48          & 48.68      & 41.12      \\
\multicolumn{1}{c|}{}                                & \multicolumn{1}{c|}{}                                                                                                 & \multicolumn{1}{c|}{(0.2, 50)} & -1.28          & 29.18          & 52.06      & 44.38      \\
\multicolumn{1}{c|}{}                                & \multicolumn{1}{c|}{}                                                                                                 & \multicolumn{1}{c|}{(0.4, 1)}  & -0.75          & 9.90           & 24.15      & 18.25      \\
\multicolumn{1}{c|}{}                                & \multicolumn{1}{c|}{}                                                                                                 & \multicolumn{1}{c|}{(0.4, 3)}  & -1.24          & 42.11          & 59.86      & 54.02      \\
\multicolumn{1}{c|}{}                                & \multicolumn{1}{c|}{}                                                                                                 & \multicolumn{1}{c|}{(0.4, 5)}  & -1.39          & 49.02          & 61.35      & 57.98      \\
\multicolumn{1}{c|}{}                                & \multicolumn{1}{c|}{}                                                                                                 & \multicolumn{1}{c|}{(0.4, 10)} & -1.61          & 55.31          & 62.70      & 59.77      \\
\multicolumn{1}{c|}{}                                & \multicolumn{1}{c|}{}                                                                                                 & \multicolumn{1}{c|}{(0.4, 15)} & -1.81          & 58.85          & 63.90      & 61.28      \\
\multicolumn{1}{c|}{}                                & \multicolumn{1}{c|}{}                                                                                                 & \multicolumn{1}{c|}{(0.4, 20)} & -1.96          & 61.62          & 63.44      & 62.15      \\
\multicolumn{1}{c|}{}                                & \multicolumn{1}{c|}{}                                                                                                 & \multicolumn{1}{c|}{(0.4, 50)} & -2.16          & 70.77          & 69.42      & 70.08      \\
\multicolumn{1}{c|}{}                                & \multicolumn{1}{c|}{}                                                                                                 & \multicolumn{1}{c|}{(0.6, 1)}  & -0.78          & 10.68          & 25.54      & 19.60      \\
\multicolumn{1}{c|}{}                                & \multicolumn{1}{c|}{}                                                                                                 & \multicolumn{1}{c|}{(0.6, 3)}  & -1.44          & 62.11          & 70.46      & 68.52      \\
\multicolumn{1}{c|}{}                                & \multicolumn{1}{c|}{}                                                                                                 & \multicolumn{1}{c|}{(0.6, 5)}  & -1.62          & 69.89          & 73.76      & 72.76      \\
\multicolumn{1}{c|}{}                                & \multicolumn{1}{c|}{}                                                                                                 & \multicolumn{1}{c|}{(0.6, 10)} & -1.99          & 76.58          & 72.12      & 73.28      \\
\multicolumn{1}{c|}{}                                & \multicolumn{1}{c|}{}                                                                                                 & \multicolumn{1}{c|}{(0.6, 15)} & -2.29          & 78.88          & 66.85      & 70.21      \\
\multicolumn{1}{c|}{}                                & \multicolumn{1}{c|}{}                                                                                                 & \multicolumn{1}{c|}{(0.6, 20)} & -2.51          & 80.03          & 59.57      & 65.73      \\
\multicolumn{1}{c|}{}                                & \multicolumn{1}{c|}{}                                                                                                 & \multicolumn{1}{c|}{(0.6, 50)} & -3.39          & 84.08          & 29.30      & 38.70      \\
\multicolumn{1}{c|}{}                                & \multicolumn{1}{c|}{}                                                                                                 & \multicolumn{1}{c|}{(0.8, 1)}  & -0.75          & 9.87           & 23.94      & 18.18      \\
\multicolumn{1}{c|}{}                                & \multicolumn{1}{c|}{}                                                                                                 & \multicolumn{1}{c|}{(0.8, 3)}  & -1.97          & 77.12          & \textbf{74.94}      & \textbf{74.07}      \\
\multicolumn{1}{c|}{}                                & \multicolumn{1}{c|}{}                                                                                                 & \multicolumn{1}{c|}{(0.8, 5)}  & -2.56          & 81.38          & 62.02      & 66.39      \\
\multicolumn{1}{c|}{}                                & \multicolumn{1}{c|}{}                                                                                                 & \multicolumn{1}{c|}{(0.8, 10)} & -3.51          & 84.52          & 29.78      & 40.97      \\
\multicolumn{1}{c|}{}                                & \multicolumn{1}{c|}{}                                                                                                 & \multicolumn{1}{c|}{(0.8, 15)} & -4.03          & 85.84          & 17.31      & 23.74      \\
\multicolumn{1}{c|}{}                                & \multicolumn{1}{c|}{}                                                                                                 & \multicolumn{1}{c|}{(0.8, 20)} & -4.34          & 86.77          & 14.52      & 18.48      \\
\multicolumn{1}{c|}{}                                & \multicolumn{1}{c|}{}                                                                                                 & \multicolumn{1}{c|}{(0.8, 50)} & -5.13          & 87.85          & 12.21      & 12.96      \\
\multicolumn{1}{c|}{}                                & \multicolumn{1}{c|}{}                                                                                                 & \multicolumn{1}{c|}{(1.0, 1)}  & -0.75          & 9.85           & 24.05      & 18.18      \\
\multicolumn{1}{c|}{}                                & \multicolumn{1}{c|}{}                                                                                                 & \multicolumn{1}{c|}{(1.0, 3)}  & -3.11          & 83.95          & 58.92      & 63.33      \\
\multicolumn{1}{c|}{}                                & \multicolumn{1}{c|}{}                                                                                                 & \multicolumn{1}{c|}{(1.0, 5)}  & -3.93          & 85.42          & 29.09      & 37.63      \\
\multicolumn{1}{c|}{}                                & \multicolumn{1}{c|}{}                                                                                                 & \multicolumn{1}{c|}{(1.0, 10)} & -4.73          & 86.96          & 13.36      & 17.28      \\
\multicolumn{1}{c|}{}                                & \multicolumn{1}{c|}{}                                                                                                 & \multicolumn{1}{c|}{(1.0, 15)} & -5.03          & 87.51          & 11.51      & 12.76      \\
\multicolumn{1}{c|}{}                                & \multicolumn{1}{c|}{}                                                                                                 & \multicolumn{1}{c|}{(1.0, 20)} & -5.20          & 87.62          & 11.31      & 11.69      \\
\multicolumn{1}{c|}{}                                & \multicolumn{1}{c|}{}                                                                                                 & \multicolumn{1}{c|}{(1.0, 50)} & -5.76          & \textbf{87.96}          & 11.28      & 10.85      \\ \cline{2-7} 
\multicolumn{1}{c|}{}                                & \multicolumn{1}{c|}{Adaptive contrastive search}                                                                      & \multicolumn{1}{c|}{-}         & -1.68          & \textbf{93.94}$\star$          & 79.61      & \textbf{85.72}$\star$      \\ 
\multicolumn{1}{c|}{}                                & \multicolumn{1}{c|}{(Hyperparameter-free)}                                                                      & \multicolumn{1}{c|}{}         &           &         &       &       \\\hline
\multicolumn{1}{c|}{\multirow{18}{*}{Stochastic}}    & \multicolumn{1}{c|}{\multirow{6}{*}{\begin{tabular}[c]{@{}c@{}}Sampling with temperature\\      ($t$)\end{tabular}}} & \multicolumn{1}{c|}{0.1}       & \textbf{-0.76}          & 10.56          & 25.19      & 19.26      \\
\multicolumn{1}{c|}{}                                & \multicolumn{1}{c|}{}                                                                                                 & \multicolumn{1}{c|}{0.3}       & -0.92          & 17.16          & 37.28      & 28.97      \\
\multicolumn{1}{c|}{}                                & \multicolumn{1}{c|}{}                                                                                                 & \multicolumn{1}{c|}{0.5}       & -1.20          & 34.88          & 60.67      & 50.63      \\
\multicolumn{1}{c|}{}                                & \multicolumn{1}{c|}{}                                                                                                 & \multicolumn{1}{c|}{0.7}       & -1.70          & 66.07          & 82.85      & 75.78      \\
\multicolumn{1}{c|}{}                                & \multicolumn{1}{c|}{}                                                                                                 & \multicolumn{1}{c|}{0.9}       & -2.48          & 88.38          & 90.85      & \textbf{84.58}      \\
\multicolumn{1}{c|}{}                                & \multicolumn{1}{c|}{}                                                                                                 & \multicolumn{1}{c|}{1}         & -3.07          & \textbf{93.27}          & \textbf{91.17}$\star$      & 83.30      \\ \cline{2-7} 
\multicolumn{1}{c|}{}                                & \multicolumn{1}{c|}{\multirow{7}{*}{\begin{tabular}[c]{@{}c@{}}Top-$k$\\      ($k$)\end{tabular}}}                        & \multicolumn{1}{c|}{1}         & \textbf{-0.76}          & 10.10          & 24.93      & 18.66      \\
\multicolumn{1}{c|}{}                                & \multicolumn{1}{c|}{}                                                                                                 & \multicolumn{1}{c|}{3}         & -1.41          & 50.04          & 72.75      & 63.86      \\
\multicolumn{1}{c|}{}                                & \multicolumn{1}{c|}{}                                                                                                 & \multicolumn{1}{c|}{5}         & -1.66          & 65.69          & 81.48      & 74.82      \\
\multicolumn{1}{c|}{}                                & \multicolumn{1}{c|}{}                                                                                                 & \multicolumn{1}{c|}{10}        & -1.90          & 77.10          & \textbf{85.80}      & \textbf{80.28}      \\
\multicolumn{1}{c|}{}                                & \multicolumn{1}{c|}{}                                                                                                 & \multicolumn{1}{c|}{15}        & -1.98          & 77.67          & 84.40      & 78.65      \\
\multicolumn{1}{c|}{}                                & \multicolumn{1}{c|}{}                                                                                                 & \multicolumn{1}{c|}{20}        & -2.05          & 79.20          & 84.21      & 78.86      \\
\multicolumn{1}{c|}{}                                & \multicolumn{1}{c|}{}                                                                                                 & \multicolumn{1}{c|}{50}        & -2.23          & \textbf{79.85}          & 85.33      & 78.32      \\ \cline{2-7} 
\multicolumn{1}{c|}{}                                & \multicolumn{1}{c|}{\multirow{5}{*}{\begin{tabular}[c]{@{}c@{}}Top-$p$ (nucleus)\\      ($p$)\end{tabular}}}              & \multicolumn{1}{c|}{0.6}       & \textbf{-1.36}          & 46.66          & 72.96      & 62.86      \\
\multicolumn{1}{c|}{}                                & \multicolumn{1}{c|}{}                                                                                                 & \multicolumn{1}{c|}{0.7}       & -1.56          & 61.38          & 80.80      & 73.60      \\
\multicolumn{1}{c|}{}                                & \multicolumn{1}{c|}{}                                                                                                 & \multicolumn{1}{c|}{0.8}       & -1.78          & 74.50          & 85.84      & 80.76      \\
\multicolumn{1}{c|}{}                                & \multicolumn{1}{c|}{}                                                                                                 & \multicolumn{1}{c|}{0.9}       & -2.05          & 84.09          & 89.18      & 84.50      \\
\multicolumn{1}{c|}{}                                & \multicolumn{1}{c|}{}                                                                                                 & \multicolumn{1}{c|}{0.95}      & -2.21          & \textbf{87.60}          & \textbf{90.05}      & \textbf{85.31}      \\ \hline
\multicolumn{1}{c|}{Reference}                       & \multicolumn{1}{c|}{Human}                                                                                            & \multicolumn{1}{c|}{-}         & -2.74          & 93.28          & 100.00     & 87.37     
\end{tabular}
    }
    \caption{Aggregated automatic evaluation results: Average scores across all seven models and all three datasets. Weighted averages are used, accounting for the different sample sizes in the datasets. The highest scores \textit{per strategy} are highlighted in \textbf{bold}, with the best results \textit{overall} additionally marked by a $\star$.}
    \label{tab:results_datasets_combined}
\end{table*}

\begin{figure*}[ht]
    \centering
    \includegraphics[width=\textwidth]{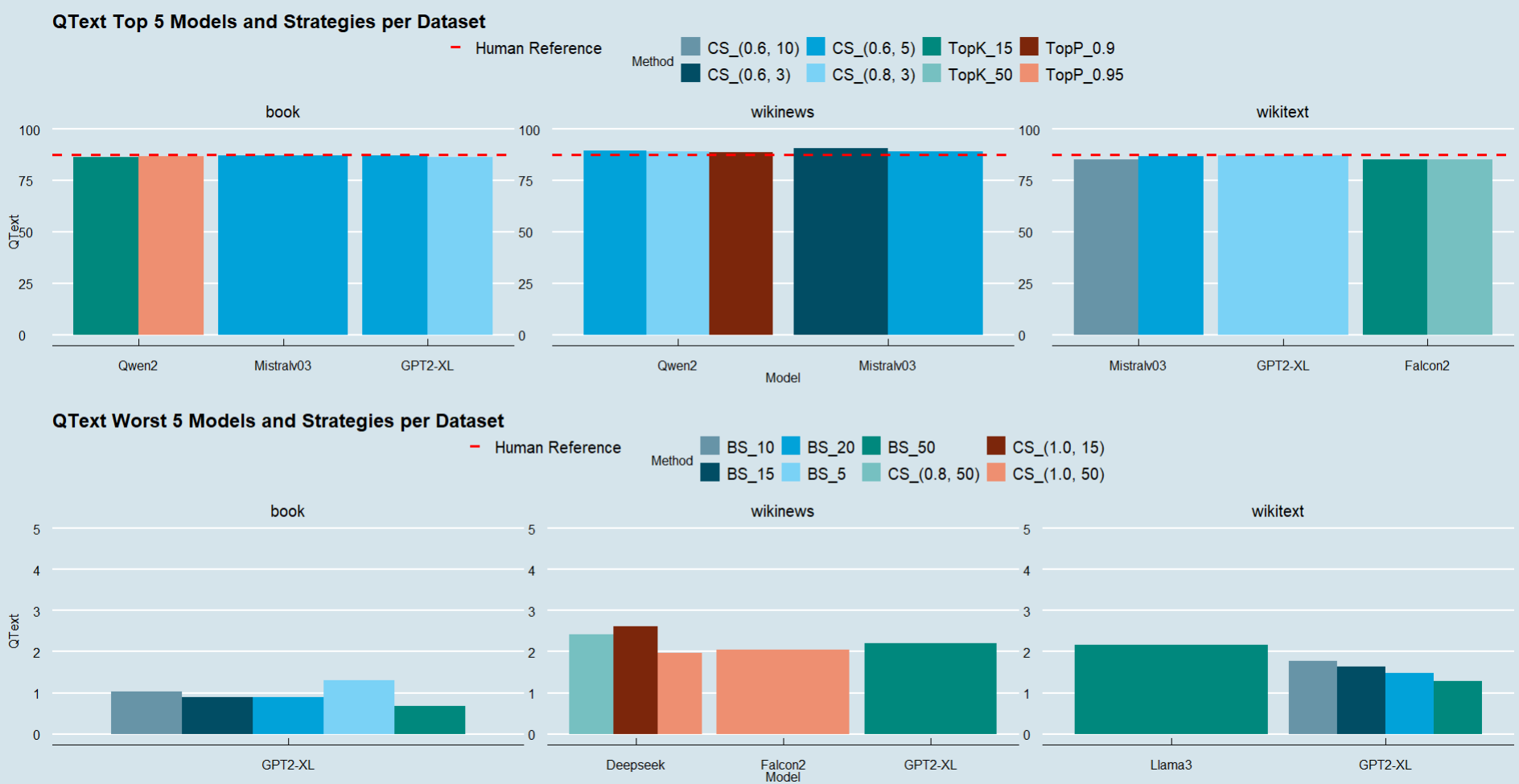}
    \caption{Top five and bottom five decoding strategies, based on QText averages for each dataset. The highest-ranking strategies generally strike a balance between coherence and diversity, while the lowest-ranking strategies tend to overemphasize one at the expense of the other—such as beam search, which favors coherence, or contrastive search with $\alpha = 1.0$ and $k = 50$, which prioritizes diversity.}
    \label{fig:qtext_best_worst}
\end{figure*}

\paragraph{Automatic Evaluation}

Below, we present a comprehensive interpretation of the results reported in Table \ref{tab:results_datasets_combined}. These findings aggregate the performance of various decoding strategies and their associated hyperparameters across multiple models and datasets, taking into account weighted averages to reflect dataset size differences. The reported metrics include coherence, diversity, MAUVE, and QText. Notably, coherence has a theoretical maximum of 0, while diversity, MAUVE, and QText each have a theoretical maximum of 100. The final row represents human-produced text, with reference values of approximately -2.74 for coherence, 93.28 for diversity, 100 for MAUVE, and 87.37 for QText, serving as an aspirational benchmark for automated methods.

Beam search yields coherence values around -0.55 to -0.72, which are far from the human coherence of -2.74. Its diversity remains low, peaking at only 12.78 and dropping to 6.17 as the beam width increases. MAUVE and QText scores under beam search never exceed 25.27 and 21.50, respectively, well below the human benchmarks.

Contrastive search achieves its best performance with moderate values of \(k\) and \(\alpha\), such as \((\alpha=0.8, k=3)\) or \((\alpha=0.6, k=10)\). However, combinations involving higher values of \(k\) and \(\alpha\) tend to reduce coherence to levels considerably below human-like values, sometimes reaching as low as -5.76. In contrast, smaller values of the degeneration penalty, such as \(\alpha=0.2\), are associated with very high coherence but extremely low diversity, resembling the performance characteristics of beam search. Even with optimal hyperparameter settings, achieving coherence and diversity levels comparable to human performance remains a challenge. 

Adaptive contrastive search (hyperparameter-free) achieves strong overall performance. It reaches a diversity of 93.94 and a QText of 85.72, both near human levels. Its coherence is -1.68, which is closer to human than many contrastive search configurations, but still not ideal. MAUVE is 79.61, which is better than basic beam search but not as high as the best stochastic approaches.

Sampling with temperature exhibits a trade-off between coherence and diversity that becomes more favorable at higher temperatures. At a temperature of 1.0, for instance, diversity reaches 93.27, MAUVE is 91.17, and QText is 83.30, all nearing human references. Coherence is -3.07, which is closer to human coherence than beam search or low-temperature sampling.

Top-\(k\) sampling improves diversity and MAUVE as \(k\) grows. At \(k=10\), diversity is 77.10, MAUVE is 85.80, and QText is 80.28, all improved over lower \(k\) values. However, coherence remains around -1.90, which is not close enough to the human target.

Top-\(p\) (nucleus) sampling offers strong performance when \(p=0.95\). At this value, diversity is 87.60, MAUVE is 90.05, and QText is 85.31, all close to human scores. Coherence at this setting is -2.21, which is closer to the human reference than many other strategies.  Figure \ref{fig:figure_one} illustrates this effect over the distribution of coherence and diversity for Mistralv03 outputs.

No single decoding strategy achieves human-level scores across all metrics; however, certain configurations exhibit performance closely aligned with human references. Notable examples include top-\(p\) sampling at \(p=0.95\), adaptive contrastive search, and sampling with temperature at \(t=1.0\), which emerge as particularly effective methods.

These findings reflect aggregated performance across three datasets and seven models. Detailed results for individual models and configurations are presented in Figures \ref{fig:bs_boxplot_model} through \ref{fig:topp_metrics_model} and discussed in Section \ref{sec:practice}. Additionally, Figure \ref{fig:qtext_best_worst} highlights the top five best- and worst-performing combinations of models and decoding strategies based on QText. A notable observation is the frequent appearance of older and smaller architectures, such as GPT2-XL (1.5B) with configurations like CS\_(0.6, 5) or CS\_(0.8, 3), among the top performers. This underscores the importance of hyperparameter selection over model size. Conversely, the lowest-performing configurations consistently overemphasized a single metric, such as beam search prioritizing coherence or contrastive search with high \(\alpha\) values emphasizing diversity. The inclusion of much larger architectures, such as Falcon2 (11B), among the lowest-performing configurations further indicates that model size alone does not guarantee superior performance. Case studies demonstrating the impact of hyperparameter choices on text generation are provided in Tables \ref{tab:case_study_1}, \ref{tab:case_study_2}, and \ref{tab:case_study_3}.

\begin{table}[!ht]
    \centering
    \resizebox{.5\textwidth}{!}{
    \begin{tabular}{l l c c c}
\toprule
\textbf{Dataset} & \textbf{Model} & \textbf{Decoding Strategy} & \textbf{Human Score} & \textbf{QText} \\
\midrule
\multicolumn{5}{c}{\textbf{Top five decoding methods per dataset}} \\
\midrule
\textbf{book} & Mistralv03 & CS\_(0.4, 10) & 97.50 & 78.31 \\
 & GPT2-XL & ACS & 97.50 & 86.03 \\
 & Mistralv03 & CS\_(0.4, 20) & 95.00 & 82.30 \\
 & Mistralv03 & TopP\_0.95 & 95.00 & 85.97 \\
 & Mistralv03 & TopP\_0.95 & 95.00 & 85.97 \\[5pt]

\textbf{wikinews} & Mistralv03 & CS\_(0.6, 10) & 97.50 & 87.48 \\
 & Mistralv03 & CS\_(0.4, 10) & 95.00 & 88.12 \\
 & Mistralv03 & TopK\_50 & 95.00 & 87.10 \\
 & GPT2-XL & ACS & 92.50 & 86.31 \\
 & Mistralv03 & TopP\_0.95 & 92.50 & 87.19 \\[5pt]

\textbf{wikitext} & GPT2-XL & ACS & 97.50 & 84.81 \\
 & Mistralv03 & CS\_(0.6, 5) & 97.50 & 86.84 \\
 & Mistralv03 & TopP\_0.95 & 97.50 & 83.22 \\
 & Mistralv03 & TopK\_50 & 95.00 & 84.89 \\
 & Mistralv03 & CS\_(0.4, 20) & 95.00 & 81.13 \\
\midrule
\multicolumn{5}{c}{\textbf{Bottom five decoding methods per dataset}} \\
\midrule
\textbf{book} & Mistralv03 & CS\_(0.8, 50) & 2.50 & 5.13 \\
 & Mistralv03 & CS\_(1.0, 50) & 2.50 & 3.10 \\
 & Mistralv03 & CS\_(0.2, 50) & 5.00 & 44.27 \\
 & Mistralv03 & CS\_(0.8, 10) & 5.00 & 9.45 \\
 & GPT2-XL & CS\_(0.8, 50) & 5.00 & 5.13 \\[5pt]

\textbf{wikinews} & Mistralv03 & BS\_20 & 2.50 & 26.08 \\
 & Mistralv03 & BS\_50 & 5.00 & 17.41 \\
 & Mistralv03 & CS\_(0.8, 50) & 5.00 & 6.73 \\
 & Mistralv03 & CS\_(1.0, 5) & 5.00 & 8.99 \\
 & Mistralv03 & CS\_(0.8, 50) & 5.00 & 6.73 \\[5pt]

\textbf{wikitext} & Mistralv03 & CS\_(1.0, 50) & 2.50 & 9.80 \\
 & Mistralv03 & CS\_(1.0, 5) & 5.00 & 29.25 \\
 & Mistralv03 & CS\_(0.8, 10) & 5.00 & 38.43 \\
 & Mistralv03 & CS\_(1.0, 50) & 5.00 & 9.80 \\
 & Mistralv03 & CS\_(0.8, 50) & 5.00 & 13.19 \\
\bottomrule
\end{tabular}
    }
    \caption{Top and bottom five models and decoding strategies for human evaluators. Results indicate a moderate, statistically significant positive correlation based on 570 evaluations, $r_{\text{Pearson}} = 0.64$ (p-value < 2.2e-16).}
    \label{tab:human_eval}
\end{table}

\paragraph{Human Evaluation}

Table \ref{tab:human_eval} summarizes the top five and bottom five decoding strategies as ranked by human evaluation scores. The findings reveal a strong alignment between humans and QText for high-performing strategies across datasets.

Specifically, in the book dataset, strategies such as ACS and CS\_(0.4, 10) achieve high human scores (97.5), which correspond closely to their QText values (86.03 and 78.31, respectively). Similarly, in the wikinews and wikitext datasets, strategies like CS\_(0.6, 10) and CS\_(0.6, 5) achieve high human ratings (97.5) and strong QText values (87.48 and 86.84, respectively). For lower-performing strategies, discrepancies between humans and QText are observed. In the book dataset, the strategy CS\_(0.2, 50) receives a low human score (5.0) but a comparatively higher QText value (44.27). Nevertheless, both human evaluations and QText consistently indicate that strategies overly prioritizing diversity, such as CS\_(1.0, 50), or coherence, such as BS\_50, tend to produce low-quality text generations.

To examine the broader alignment between automatic metrics with human preferences, we analyze their correlation. Figure \ref{fig:correlations_auto_human} illustrates that human scores have moderate, statistically significant positive correlations with QText and MAUVE. Coherence demonstrates a lower but statistically significant positive correlation, while diversity exhibits a very weak and non-significant correlation. These findings highlight limitations of automatic metrics, consistent with previous work by \citet{su2023contrastive} and \citet{arias2024adaptivecontrastivesearchuncertaintyguided}.

\section{Practical Recommendations}
\label{sec:practice}

Based on the results of our study, we provide the following practical recommendations for selecting decoding strategies in open-ended text generation. 

\paragraph{Deterministic Strategies}

Beam search is generally not recommended for open-ended text generation tasks due to its propensity to produce outputs that fail to capture the diversity of human language. As illustrated in Figure \ref{fig:bs_metrics_model}, our findings demonstrate consistently low performance across all models and beam width settings when compared to human references across four evaluation metrics. A notable trend is that smaller beam widths (e.g. 3 or 5) tend to yield better results; however, even at these settings, the performance disparity remains substantial, particularly concerning MAUVE scores and diversity measures. While the Falcon2 (11B) model emerged as the best performer in this task, it still underperforms across all four metrics.

\paragraph{Sampling Strategies}

Sampling with temperature is highly sensitive to its temperature hyperparameter. Figure \ref{fig:swt_metrics_model} illustrates that performance increases monotonically for all models and metrics as temperature rises in \( t \in [0.1, 1.0] \). The best results are consistently observed at temperatures of 0.9 or 1.0. Both Falcon2 and Qwen2 perform best overall, though all models achieve strong results at higher temperatures. Larger models do not show a clear advantage over smaller ones. Specifically, GPT2-XL (1.5B) might outperform Llama3 (8B) under the same hyperparameter configuration.

For top-\( k \) sampling, performance is also sensitive to the truncation hyperparameter \( k \). As shown in Figure \ref{fig:topk_metrics_model}, small \( k \) values (e.g., \( k = 3, 5 \)) produce lower performance, while medium to larger \( k \) values (e.g., \( k = 10, 15, 20, 50 \)) yield higher performance. The highest results occur at \( k = 20 \) or \( k = 50 \), and performance similar to human references is achievable with \( k = 10 \) or \( k = 15 \). Qwen2 reaches the best overall performance, but most models perform well at higher \( k \) values. Llama3 is an exception, showing inconsistent performance and lower scores at higher \( k \) values. Similar to the temperature results, larger models do not reliably outperform smaller ones. Specifically, GPT2-XL (1.5B) performs as well as or better than Llama3 (8B), and Qwen2 (7B) outperforms Falcon2 (11B).

For top-\( p \) (nucleus) sampling, Figure \ref{fig:topp_metrics_model} shows that for all models, performance increases monotonically as \( p \) moves from 0.6 to 0.95. The best results occur at \( p = 0.9 \) or \( p = 0.95 \), consistent with previous work by \citet{holtzman2019curious}. Falcon2 and Qwen2 achieve the highest overall performance, but all models perform well at these higher values. As with the other sampling strategies, larger models do not consistently dominate smaller ones.

It is worth mentioning that combinations of various stochastic decoding strategies are possible, e.g. first adjusting temperature values before truncation with hyperparameters $k$ or $p$. However, this type of analysis lies beyond the scope of the present work.

\paragraph{Contrastive Strategies}

Figures \ref{fig:cs_metrics_deepseek} through \ref{fig:csb_metrics_qwen} illustrate that optimal Contrastive Search performance depends on balancing the hyperparameters \( \alpha \) (degeneration penalty) and \( k \) (truncation length). The best overall combination was observed at \( \alpha = 0.6 \) and \( k = 5 \), which aligns with previous work by \citet{su2022contrastive}. Other settings of  \((\alpha, k)\) such as \((0.8, 3) \), \( (0.4, 20) \), and \( (0.4, 50) \), also maintained a suitable trade-off between diversity and coherence, performing similar to human references. Very low and very high values of \( \alpha \) reduced performance, regardless of the choice of $k$. Specifically, \( \alpha = 0.2 \) produced results similar to beam search, with low diversity and high coherence. In contrast, \( \alpha = 1.0 \) generated outputs with high diversity but very low coherence. These findings align with the work by \citet{su2022empirical}. Under balanced conditions, all models except Llama3 approached human-level performance. The Llama3 model favored coherence but showed limited diversity, indicating degeneration and suggesting that it may be unsuitable for open-ended generation with this decoding strategy.

We also evaluated the hyperparameter-free Adaptive Contrastive Search. Figure \ref{fig:acs_metrics_model} shows that although its MAUVE score is moderate, it achieves a balance of diversity and coherence that produces QText values closer to human references. As indicated in Table \ref{tab:results_datasets_combined}, it attains the highest diversity (93.95\%) and the highest QText score (87.37\%) among all tested decoding strategies and hyperparameter combinations. However, its outputs tend to be more coherent than those produced by humans, a pattern observed in most of the examined hyperparameter configurations.

\paragraph{General recommendations}

It is important to note that relying solely on automatic evaluation metrics may lead to incomplete assessments of text quality. The results presented here indicate that current automatic metrics do not fully capture human preferences. Previous studies \citep{arias2024adaptivecontrastivesearchuncertaintyguided} have shown that certain decoding methods, such as CS-based \textit{DoubleExp}, achieve high scores on metrics like MAUVE but are consistently rejected by human evaluators.

Additionally, stylistic diversity should be considered when choosing decoding strategies. For example, creative writing tasks may benefit from strategies encouraging diversity and moderate coherence, supporting narrative variation. In contrast, tasks focused on factual accuracy and coherence, such as Wikipedia or news generation (as in Wikinews), may require tighter control over coherence. Section \ref{sec:appendix} provides detailed performance analyses across different settings, offering further guidance on which strategies yield the best results for each dataset type.


\section{Conclusion}
\label{sec:conclusion}

Decoding strategies for large language models are a crucial yet often underexamined aspect of open-ended text generation. This study presents a detailed analysis of how hyperparameter selection across various decoding strategies substantially influences the quality of generated text. The findings highlight the necessity of maintaining a balance between coherence and diversity, as strategies that heavily prioritize one tend to underperform overall. Through extensive sensitivity analysis, we show that the choice of decoding method and its associated hyperparameters can impact text quality as much as, if not more than, model size.

We provide practical guidelines for decoding strategies, recommending balanced approaches such as contrastive search with moderate values of $\alpha$ and $k$, alongside hyperparameter-free methods like adaptive contrastive search. While sampling methods, such as top-$k$ and nucleus sampling, can produce high-quality text, they exhibit greater variability and demand careful tuning.

Human evaluations demonstrate a moderate correlation with automatic metrics, pointing to the need for more reliable and robust evaluation tools. Current metrics sometimes prioritize diversity over coherence, which may not always align with human judgment. Our findings support the growing consensus that more nuanced metrics are required to better capture the trade-offs considered by human evaluators when assessing text quality. These insights aim to assist practitioners in understanding the critical role of hyperparameter selection in open-ended text generation.

\section*{Limitations}

While this study provides insights into how hyperparameters affect the quality of generated text, it also has limitations. First, the impact of our findings on models that have undergone supervised fine-tuning (SFT) remains unexplored. Future work could investigate how these models perform under similar experimental conditions.

Moreover, the analysis was conducted exclusively in English, leaving the generalizability to multilingual settings untested. Previous studies \cite{su2023contrastive, arias2024adaptivecontrastivesearchuncertaintyguided} suggest that these findings may generalize to other languages, but further investigation is required.

Finally, this study focuses on analyzing the sensitivity of commonly used automatic metrics, which are inherently influenced by the models employed in the evaluation process. For example, coherence is assessed using the OPT 2.7B model \citep{zhang2022optopenpretrainedtransformer}, and MAUVE is computed using GPT-2-based embeddings. The potential biases introduced by these model choices have not been fully explored, and addressing this in future research may offer valuable insights.


\section*{Ethics Statement}

We affirm that our research adheres to the \href{https://www.aclweb.org/portal/content/acl-code-ethics}{ACL Ethics Policy}. This work involves the use of publicly available datasets and does not include any personally identifiable information. For our human evaluation, we employed third-party evaluators, ensuring a rate of over \$20 per hour. An ethical concern worth mentioning is the use of language models for text generation, which may produce harmful content, either through intentional misuse by users or unintentionally due to the training data or algorithms. We declare that there are no conflicts of interest that could potentially influence the outcomes, interpretations, or conclusions of this research. All funding sources supporting this study are acknowledged in the acknowledgments section. We have diligently documented our methodology, experiments, and results, and commit to sharing our code, data, and other relevant resources to enhance reproducibility and further advancements in the field.

\section*{Acknowledgments}

We would like to extend our gratitude to David Garces Arias for his assistance in the Human Evaluation design. Matthias Aßenmacher was funded by the Deutsche Forschungsgemeinschaft (DFG, German Research Foundation) under the National Research Data Infrastructure – NFDI 27/1 - 460037581.




\clearpage

\appendix

\onecolumn

\section{Appendix}
\label{sec:appendix}

\subsection{Case Studies}
\label{appendix:case_studies}

\begin{table}[H] 
\centering
\resizebox{1\textwidth}{!}{
\begin{tabular}{|p{4cm}|p{16cm}|}
\hline
\textbf{Prompt} & \textbf{"The atmosphere dripped with emotion. Even Sola, the 13-year old, caught the current. Her eyes, fixated on the mother-son scene",} \\ \hline
Beam search (10) & "in front of her, welled up with tears. \newline \newline 'I’m so proud of you,' the mother said. \newline \newline 'I’m so proud of you,' the son said. \newline \newline 'I’m so proud of you,' the mother said. \newline \newline 'I’m so proud of you,' the son said... (text continues)" \\ \hline
Contrastive search (0.6, 10) & "unfavor- \newline \newline ablc: 0.5pt; \newline \newline \textbackslash baselineskip=1pt ... (text continues)" \\ \hline
Contrastive search (1.0, 50) & ", were moist. \newline \newline 'I’m sorry, Mama,' the boy said. \newline \newline 'I’m sorry too, son,' the mother replied. \newline \newline 'I’m sorry for not being a good son. I’m sorry for not being there for you... (text continues)" \\ \hline
Adaptive contrastive search & ", shimmered with tears. A few minutes later, I was on the roof of the school, peering through binoculars to see what was happening down below. In the distance, I could see the silhouettes of a number of school buses, their windows blackened by fire... (text continues)" \\ \hline
Sampling with temperature (1) & "unfolding before her, darted from one to the other, but she could not see it. Not really. \newline \newline 'You’re not my son,' the mother spoke, her voice strong in spite of the tears rolling down her cheeks... (text continues)" \\ \hline
Top-k (50) & "reflected a desire to be a mother. Her heart was full of love for her mother. She felt the mother’s pain. She had lost her own mother to cancer. \newline \newline A 13-year-old, a 24-year-old, and a 40-year-old stood in the center of the church... (text continues)" \\ \hline
Top-p (0.9) & "were shining with silent approval. She had seen something unusual. She had witnessed an act of parental care. The son was trying to calm his mother. He spoke words of assurance; he told her things would work out... (text continues)" \\ \hline
Human & ", were wide with wonder. \newline \newline 'Mommy, why is he crying?' she asked softly. \newline \newline 'It’s okay, Sola. It’s okay. Don’t worry,' her mother replied... (text continues)" \\ \hline
\end{tabular}
}
\caption{Case Study: Comparison of multiple decoding strategies for a prompt taken from the Book corpus. Beam search exhibits a repetitive behavior, while contrastive search generates erratic continuations for some of the hyperparameter choices.}
\label{tab:case_study_1}
\end{table}

\begin{table}[H]
\centering
\resizebox{1\textwidth}{!}{
\begin{tabular}{|p{4cm}|p{16cm}|}
\hline
\textbf{Prompt} & \textbf{"Two French journalists kidnapped on Wednesday north-east of the capital Kabul are believed to be in good health, although the condition of their three Afghan assistants is unknown.",
} \\ \hline
Beam search (10) & "Two French journalists kidnapped on Wednesday north-east of the capital Kabul are believed to be in good health, although the condition of their three Afghan assistants is unknown. \newline \newline Two French journalists kidnapped on Wednesday north-east of the capital Kabul are believed to be in good health, although the condition of their three Afghan assistants is unknown... (text continues)" \\ \hline
Contrastive search (0.6, 10) & "Sylvain Dufoix and Nicolas Hénin of Canal+ were abducted by gunmen in broad daylight, along a road that connects Taloqan, the capital of Takhar province, 200 kilometers (125 miles). A spokesman for President Hamid Karzai condemned the 'brutality and barbarianism of terrorists who target civilians.' ... (text continues)" \\ \hline
Contrastive search (1.0, 50) & "Reporters without Borders (RSF) condemns the 'barbaric' abduction of the two reporters and their Afghan assistants, and calls on the Taliban to release them immediately. \newline \newline 'Two years after the murder of seven French journalists in Mali and five years after the killing of the American journalist James Foley by IS, the world’s media are once again being targeted with this barbaric kidnapping.' ... (text continues)" \\ \hline
Adaptive contrastive search & "French President Nicolas Sarkozy said on Thursday that he had been in contact with the Taliban, who have claimed responsibility for the kidnapping. \newline \newline 'I have asked the Taliban to release the hostages,' Sarkozy said. 'I have asked the Taliban to release the hostages unharmed.' ... (text continues)" \\ \hline
Sampling with temperature (1) & "French President Nicolas Sarkozy has confirmed that two French journalists kidnapped in northern Afghanistan were in good health, although he rejected the Taliban's call for the French government to withdraw its troops from Afghanistan. \newline \newline 'They are in good health,' Sarkozy said, speaking in New Delhi. 'We are doing everything to ensure their return.' ... (text continues)" \\ \hline
Top-k (50) & "The two French journalists, Nicolas Hénin and Edouard Elias, were kidnapped by gunmen in the Tagab district of Kapisa province, north-east of Kabul, on Wednesday. \newline \newline Their Afghan assistants, who were also kidnapped, are believed to be in poor health, according to a statement from the French foreign ministry. ... (text continues)" \\ \hline
Top-p (0.9) & "The two French journalists, Nicolas Hénin and Edouard Elias, were kidnapped by gunmen in the Tagab district of Kapisa province, north-east of Kabul, on Wednesday. \newline \newline Their Afghan assistants, who were also kidnapped, are believed to be in poor health, according to a statement from the French foreign ministry. ... (text continues)" \\ \hline
Human & "France’s foreign ministry said in a statement on Thursday that the three had been 'kidnapped by armed individuals who took them from their hotel in Herat province at gunpoint.' Afghan President Ashraf Ghani’s spokesman Aimal Faizi said a helicopter had been sent to pick up the journalists ... (text continues)" \\ \hline
\end{tabular}

}
\caption{Case Study: Comparison of multiple decoding strategies for a prompt taken from the Wikinews corpus. Beam search exhibits a repetitive behavior.}
\label{tab:case_study_2}
\end{table}


\begin{table}[H]
\centering
\resizebox{1\textwidth}{!}{
\begin{tabular}{|p{4cm}|p{16cm}|}
\hline
\textbf{Prompt} & \textbf{"Headlam served as Officer Commanding North @-@ Western Area in 1946, and as Director of Training from 1947 to 1950. In 1950 \u2013 51,",} \\ \hline
Beam search (10) & "he was Commandant of the Royal Military Academy Sandhurst. From 1951 to 1953, he was Deputy Chief of the Imperial General Staff. From 1953 to 1955, he was Chief of the Imperial General Staff... (text continues)" \\ \hline
Contrastive search (0.6, 10) & "Headlamexercised command responsi" \\ \hline
Contrastive search (1.0, 50) & "he was the Officer Commanding the 1st Battalion, The Royal Ulster Rifles. He was promoted to Brigadier in 1951, and served as the Director of Army Education from 1951 to 1954... (text continues)" \\ \hline
Adaptive contrastive search & "he was Commandant of the Royal Military Academy, Sandhurst, and from 1951 to 1953, he was Deputy Chief of the Imperial General Staff. From 1953 to 1955, he was Chief of the Imperial General Staff... (text continues)" \\ \hline
Sampling with temperature (1) & "he served as Commander, 1st Battalion, Royal Canadian Regiment. In 1951–53, he was Chief of Staff of Canadian Army Staff, and in 1953–55, he was responsible for the reorganization of the Canadian Army Staff as Director of Operations... (text continues)" \\ \hline
Top-k (50) & "he served as the Commanding Officer of the 1st Battalion of the Nigerian Regiment. Headlam was a member of the first intake of the Nigerian Military School in 1933. He attended the Royal Military College, Sandhurst in 1935... (text continues)" \\ \hline
Top-p (0.9) & "he was Commandant of Depot @-@ Training, and the next year was Deputy Chief of @-@ Staff Training. Headlam returned to the Far East, serving as Commander, Australian Troops on the Malay Peninsula from 1952 to 1954... (text continues)" \\ \hline
Human & "he was Director of Operations and Intelligence, and in 1951–54, Commander of the 1st Division, which was the most powerful division in the world. He was appointed Commander-in-Chief of the Army in 1954... (text continues)" \\ \hline
\end{tabular}

}
\caption{Case Study: Comparison of multiple decoding strategies for a prompt taken from the Wikitext corpus. Beam search exhibits a repetitive behavior, while contrastive search and sampling-based strategies generate erratic continuations for some of the hyperparameter choices.}
\label{tab:case_study_3}
\end{table}

\clearpage

\subsection{Performance per Model, Dataset and Decoding Strategy}
\label{appendix:performance_analysis}

\textit{Remark:} In the following, we exclude Mistralv01 for visualization purposes, as its behavior closely mirrors that of Mistralv03, which employs the same architecture but was trained with additional data.
\begin{figure}[H]
\centering
\includegraphics[width=1.0\textwidth]
{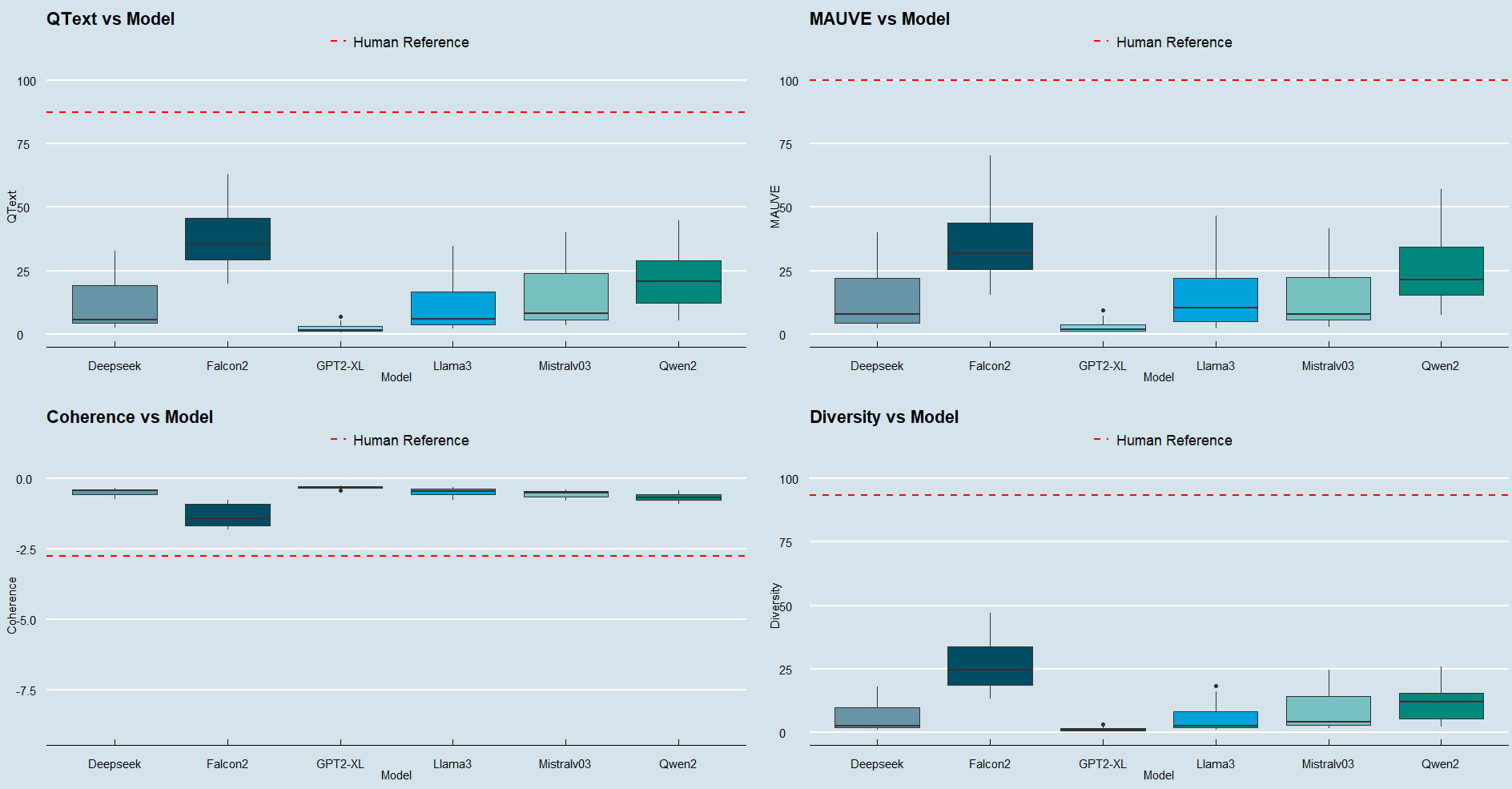}

\caption{Distribution of metric values per model, by using a \textit{Beam Search} decoding strategy.}
\label{fig:bs_boxplot_model}
\end{figure}

\begin{figure}[H]
\centering
\includegraphics[width=1.0\textwidth]
{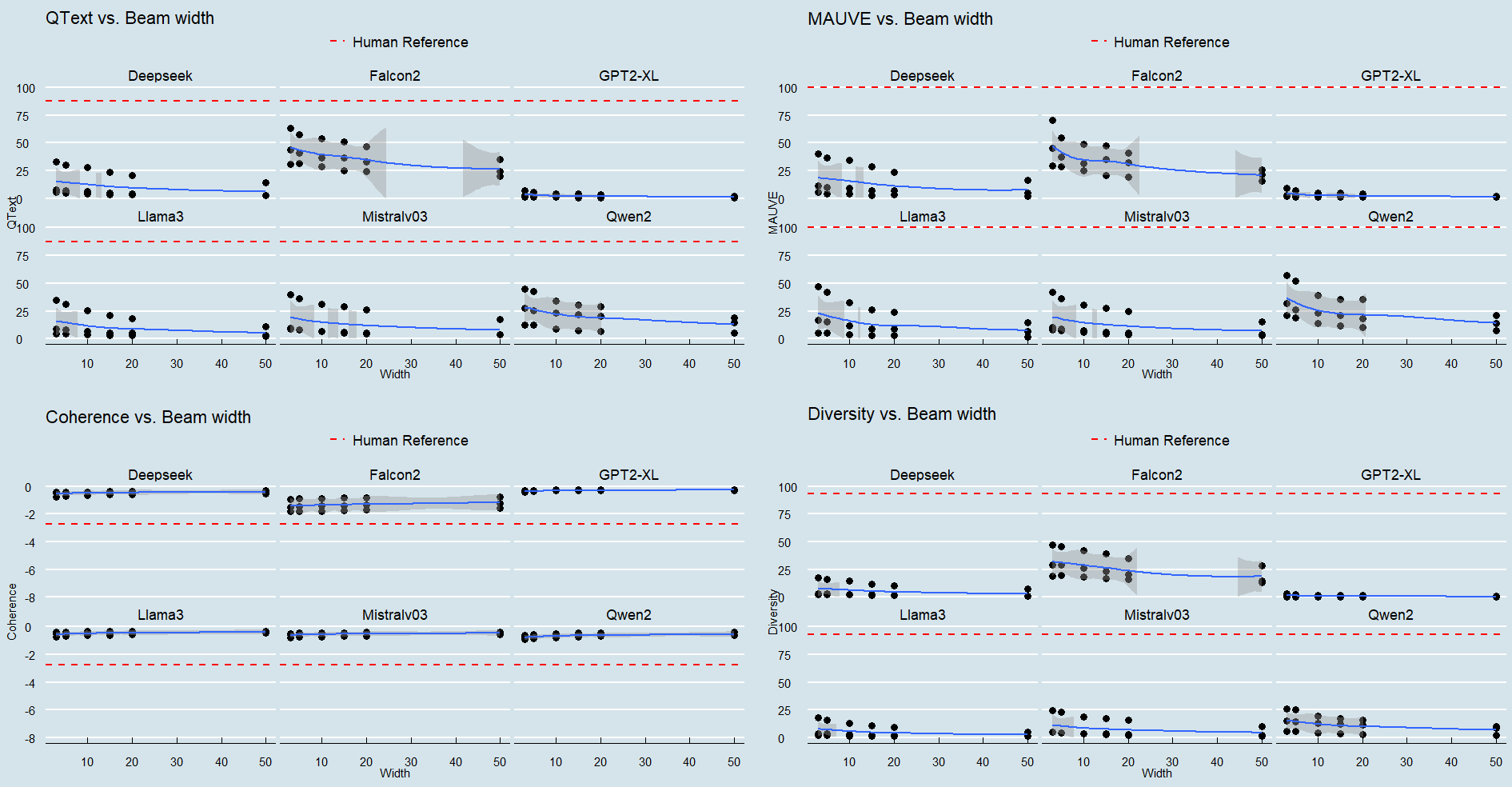}

\caption{Effect of beam width on metric behavior.}
\label{fig:bs_metrics_model}
\end{figure}

\begin{figure}[H]
\centering
\includegraphics[width=1.0\textwidth]
{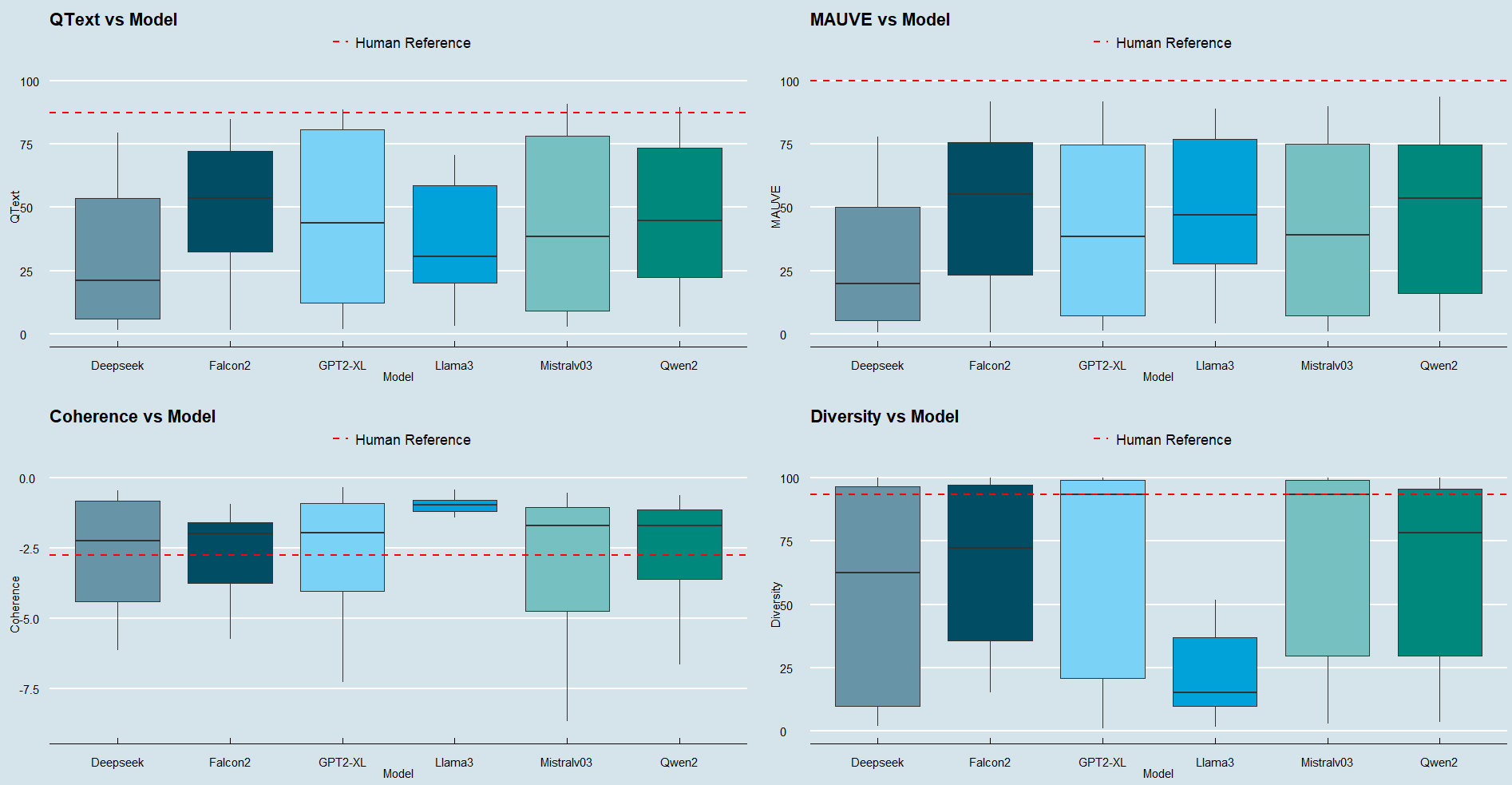}

\caption{Distribution of metric values per model, by using a \textit{Contrastive Search} decoding strategy.}
\label{fig:cs_boxplot_model}
\end{figure}

\begin{figure}[H]
\centering
\includegraphics[width=1.0\textwidth]
{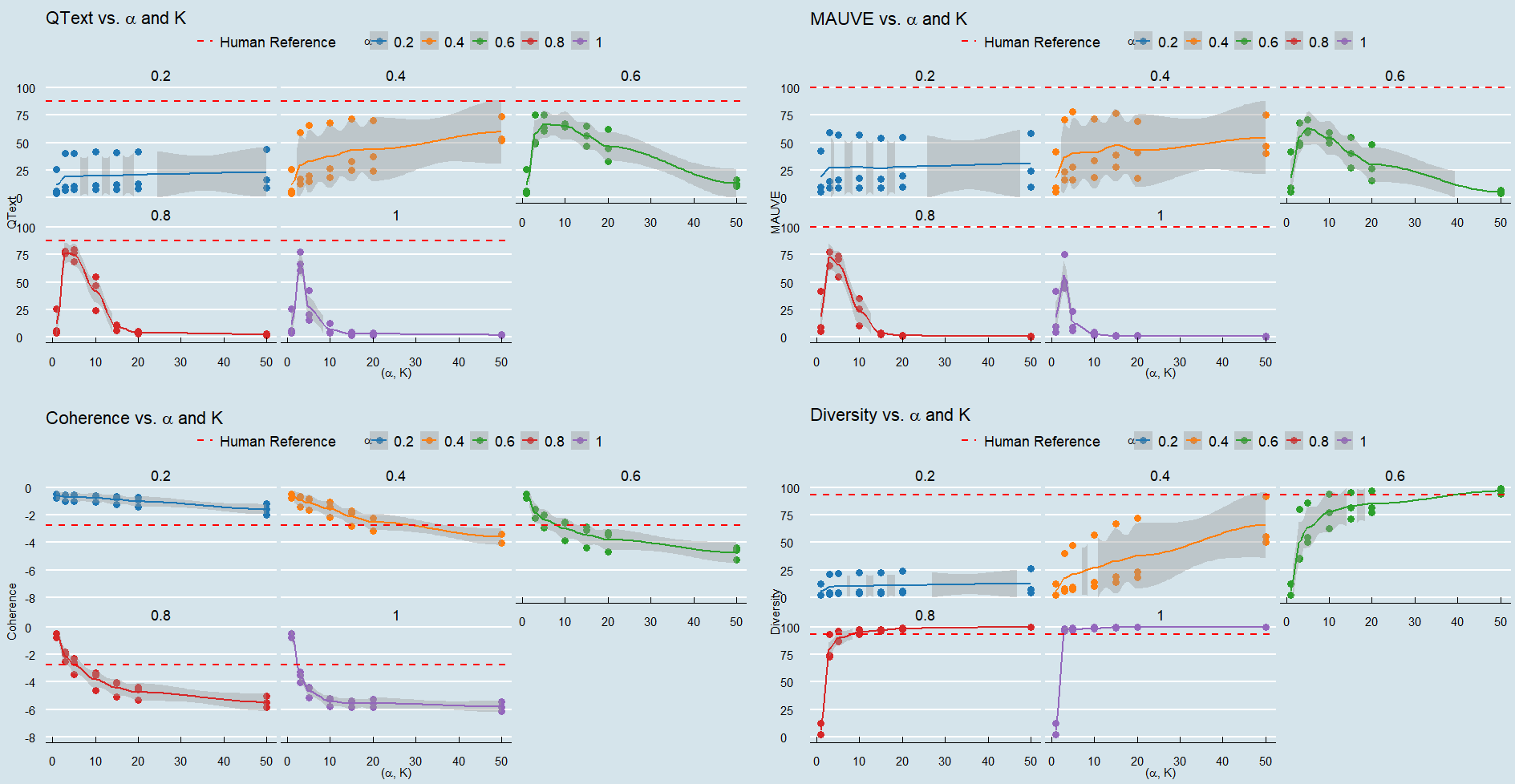}

\caption{Effect of $\alpha$ (fix) and $k$ on metric behavior (model Deepseek visualized).}
\label{fig:cs_metrics_deepseek}
\end{figure}

\begin{figure}[H]
\centering
\includegraphics[width=1.0\textwidth]
{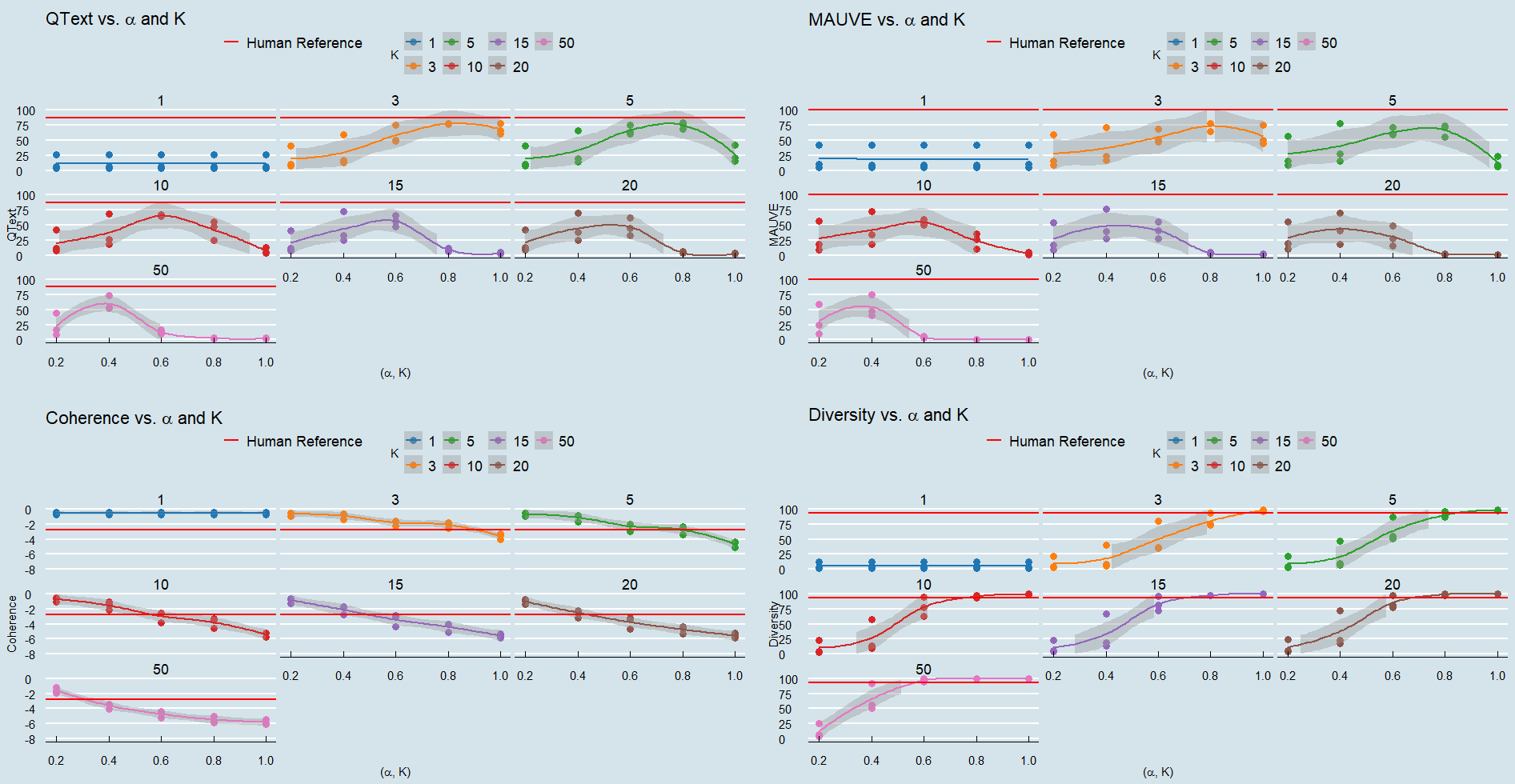}

\caption{Effect of $k$ (fix) and $\alpha$ on metric behavior (model Deepseek visualized).}
\label{fig:csb_metrics_deepseek}
\end{figure}

\begin{figure}[H]
\centering
\includegraphics[width=1.0\textwidth]
{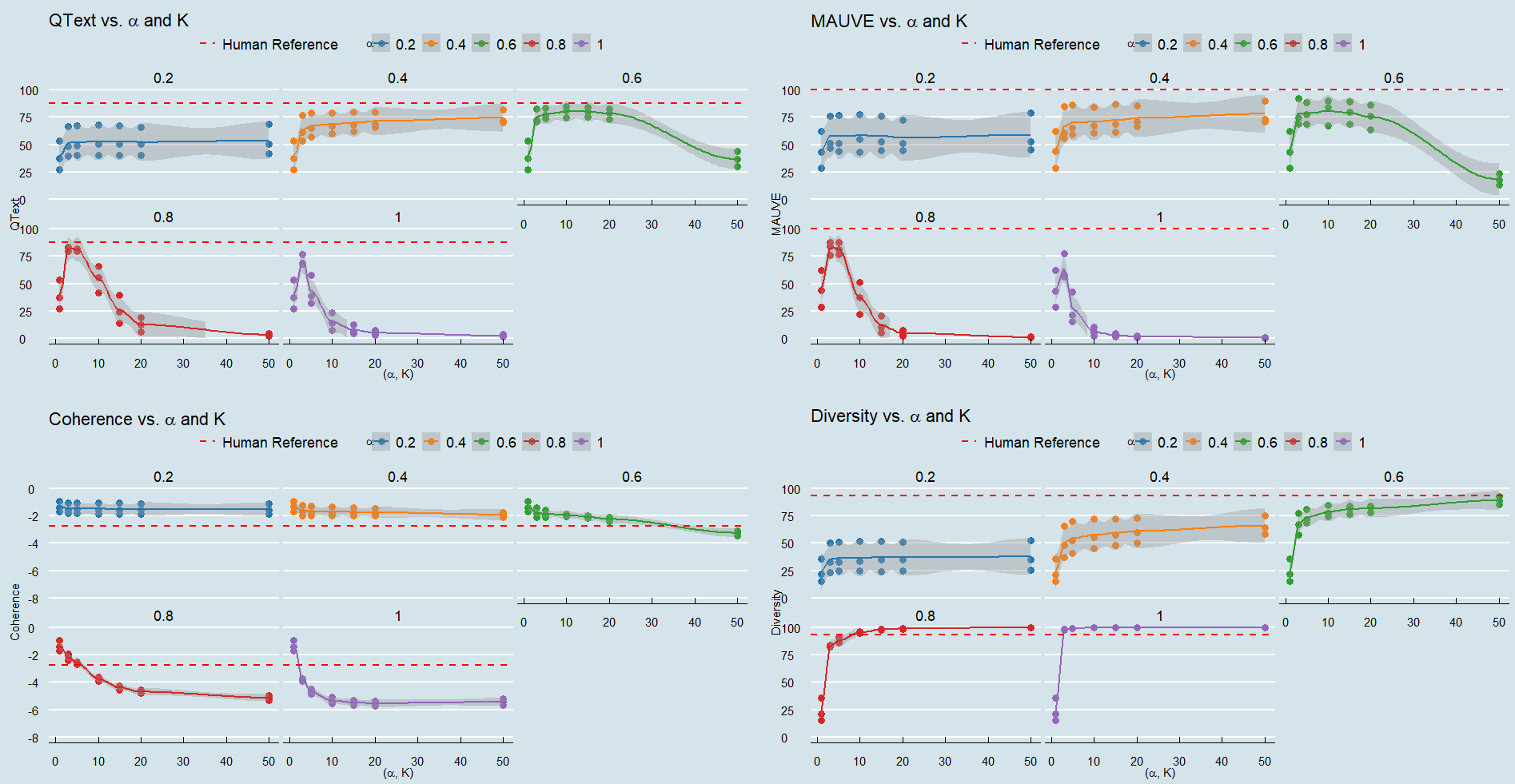}

\caption{Effect of $\alpha$ (fix) and $k$ on metric behavior (model Falcon2 visualized).}
\label{fig:cs_metrics_falcon}
\end{figure}

\begin{figure}[H]
\centering
\includegraphics[width=1.0\textwidth]
{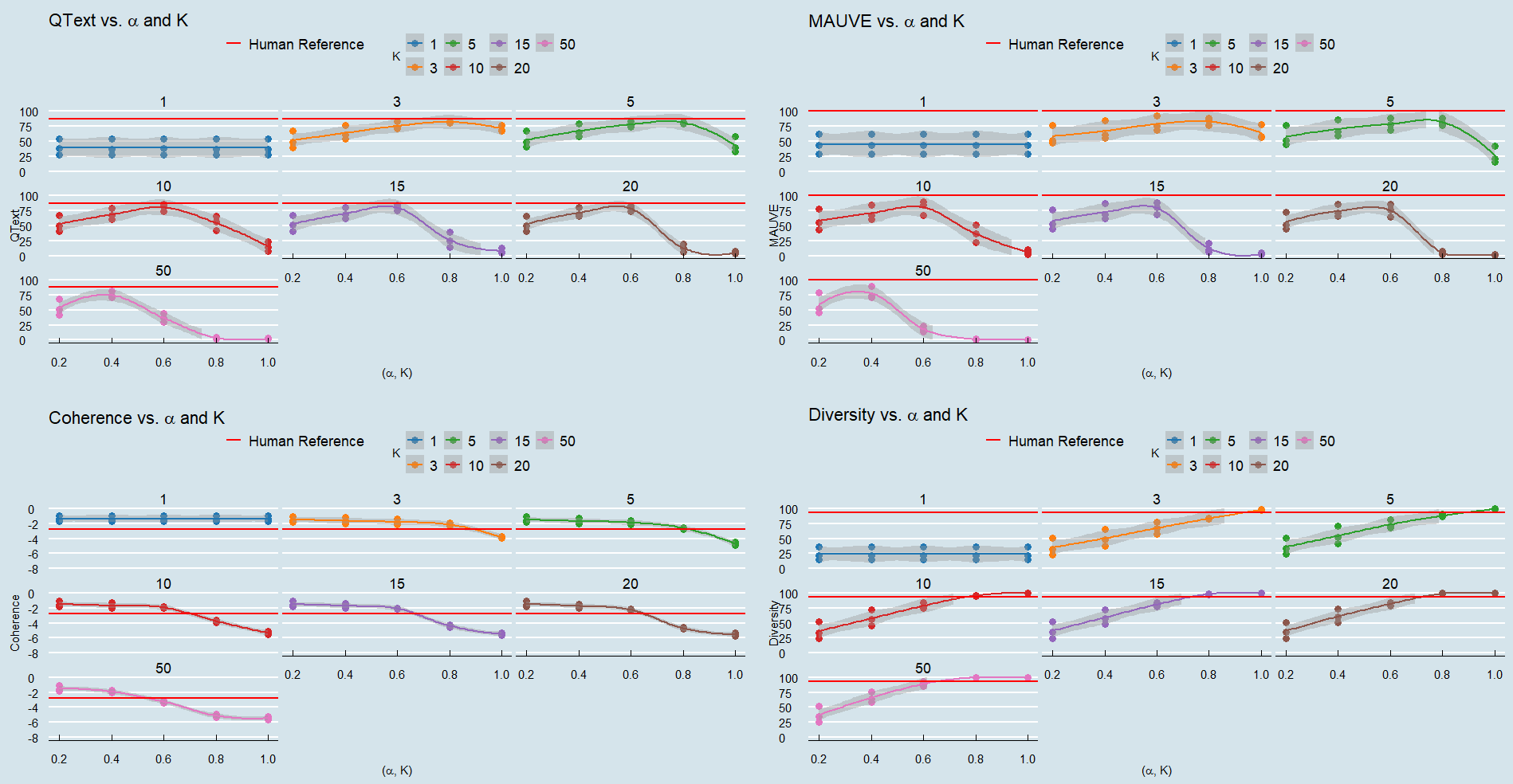}

\caption{Effect of $k$ (fix) and $\alpha$ on metric behavior (model Falcon2 visualized).}
\label{fig:csb_metrics_falcon}
\end{figure}

\begin{figure}[H]
\centering
\includegraphics[width=1.0\textwidth]
{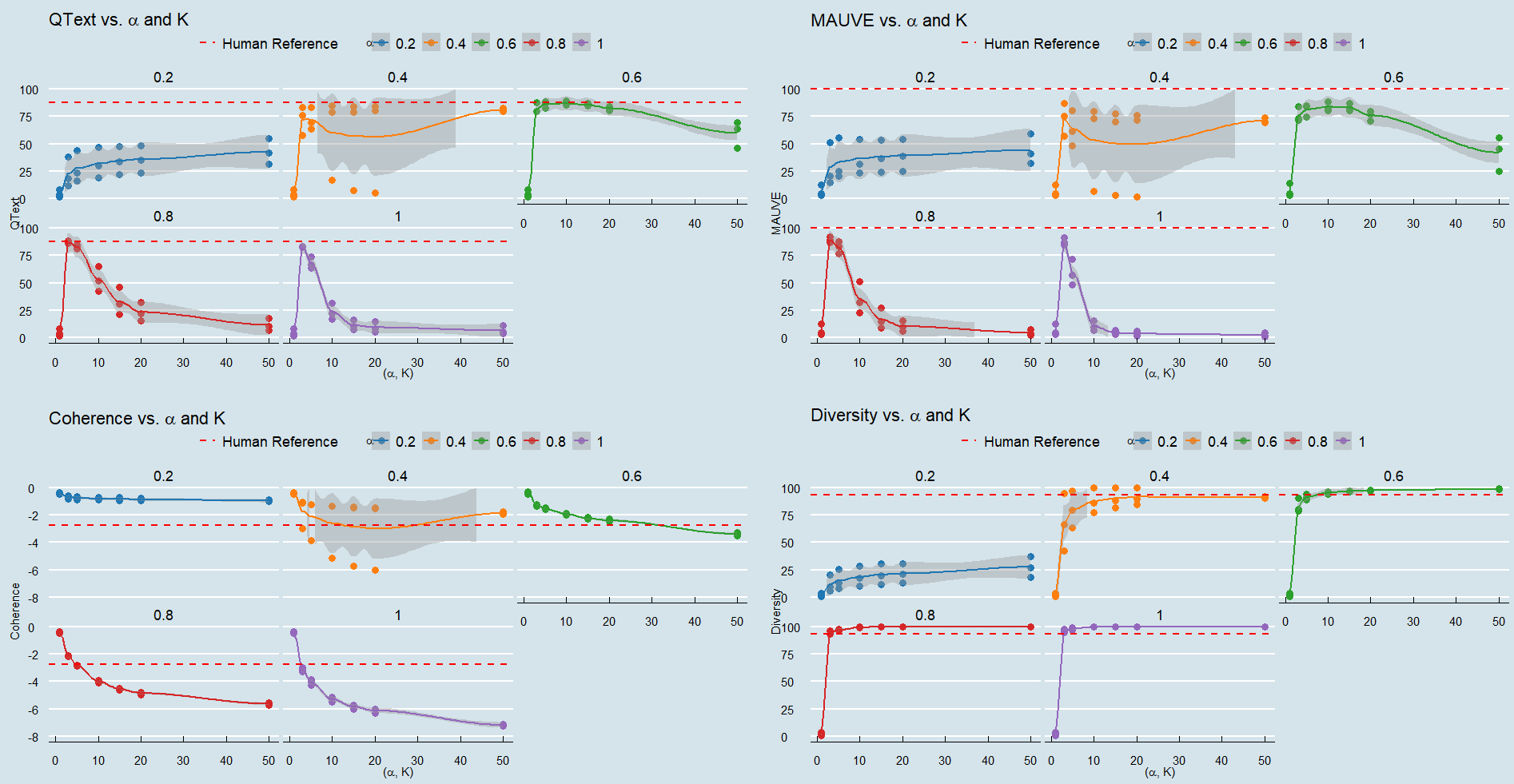}

\caption{Effect of $\alpha$ (fix) and $k$ on metric behavior (model GPT2-XL visualized).}
\label{fig:cs_metrics_gpt2xl}
\end{figure}

\begin{figure}[H]
\centering
\includegraphics[width=1.0\textwidth]
{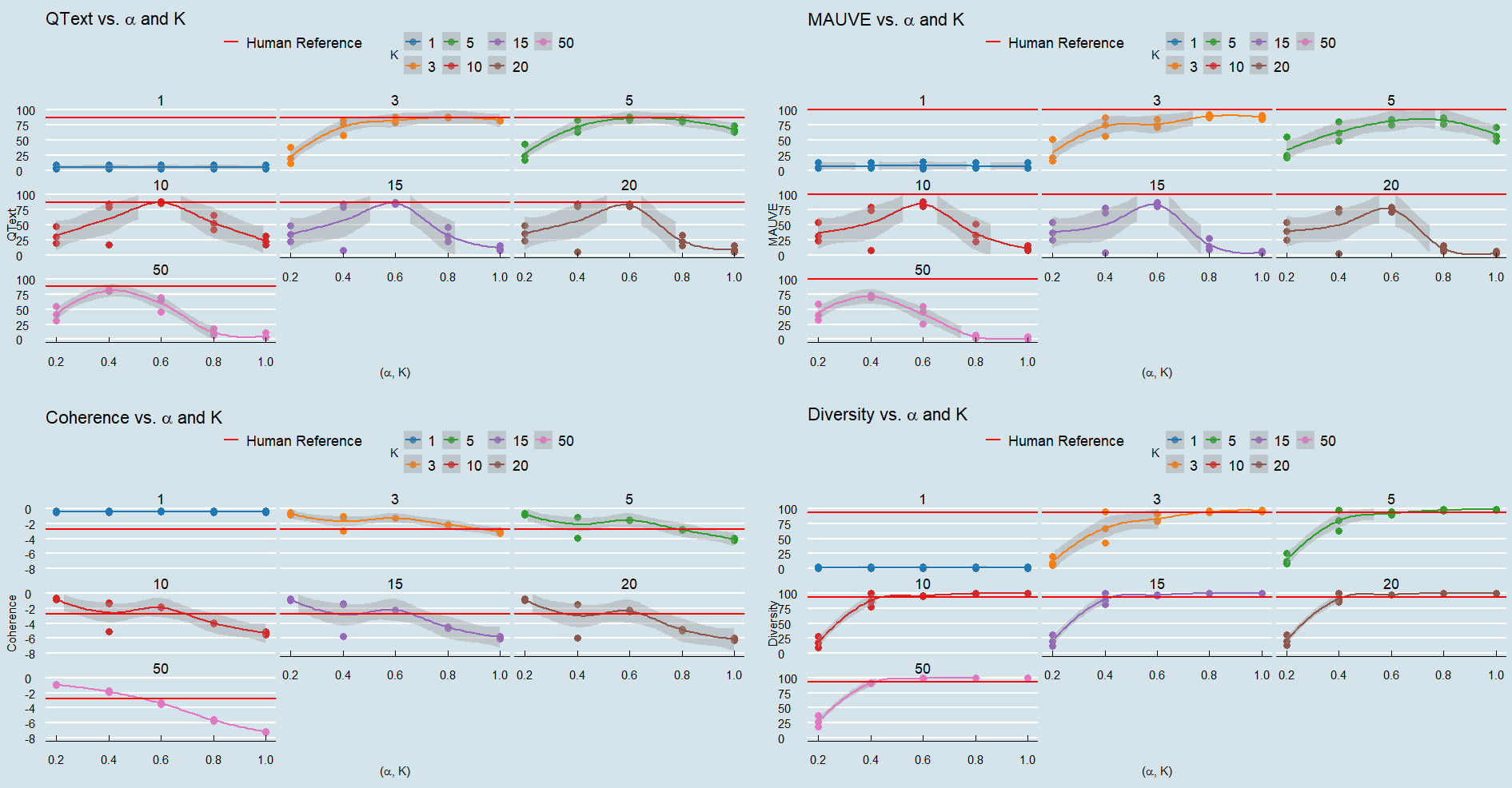}

\caption{Effect of $k$ (fix) and $\alpha$ on metric behavior (model GPT2-XL visualized).}
\label{fig:csb_metrics_gpt2xl}
\end{figure}

\begin{figure}[H]
\centering
\includegraphics[width=1.0\textwidth]
{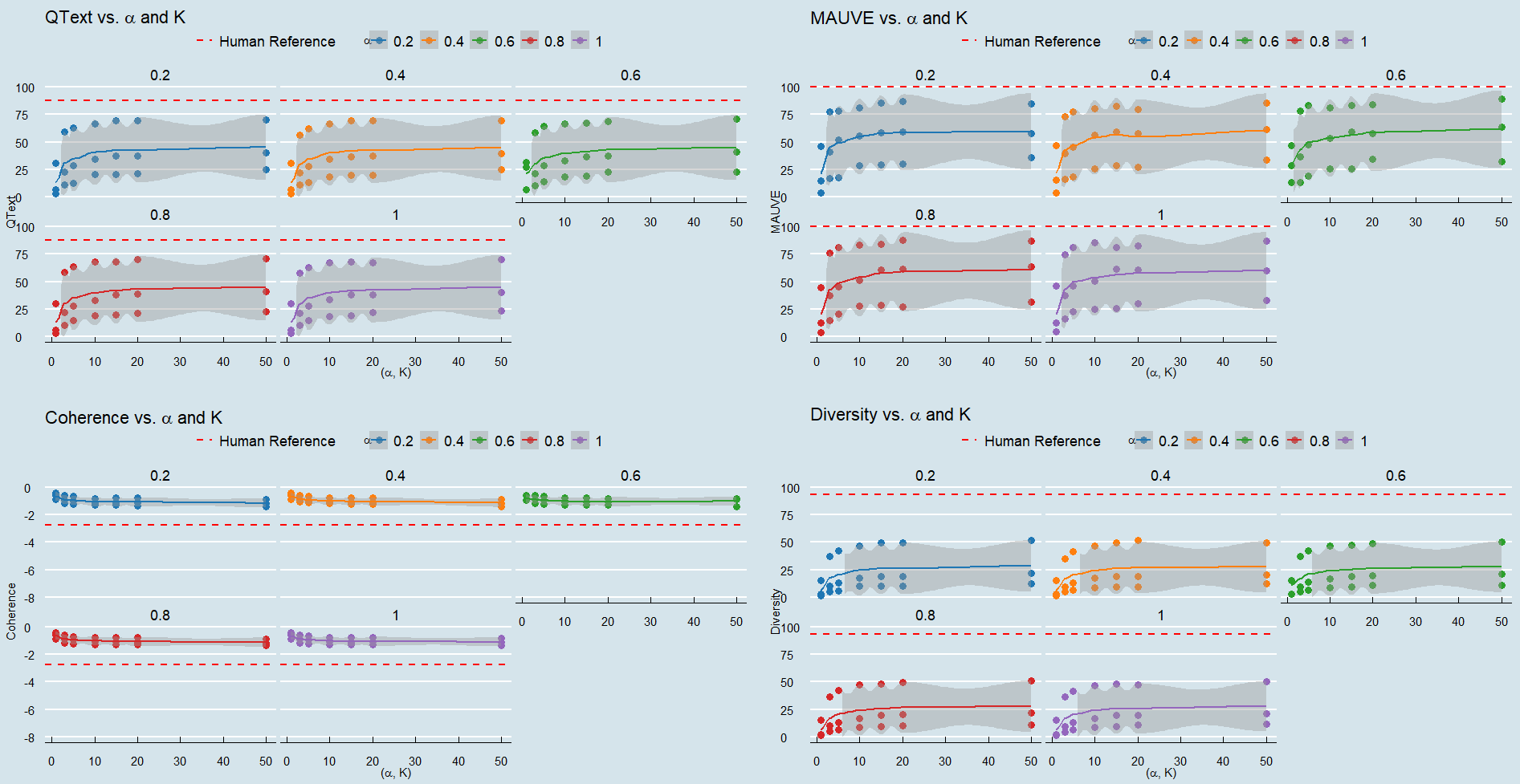}

\caption{Effect of $\alpha$ (fix) and $k$ on metric behavior (model Llama3 visualized).}
\label{fig:cs_metrics_llama}
\end{figure}

\begin{figure}[H]
\centering
\includegraphics[width=1.0\textwidth]
{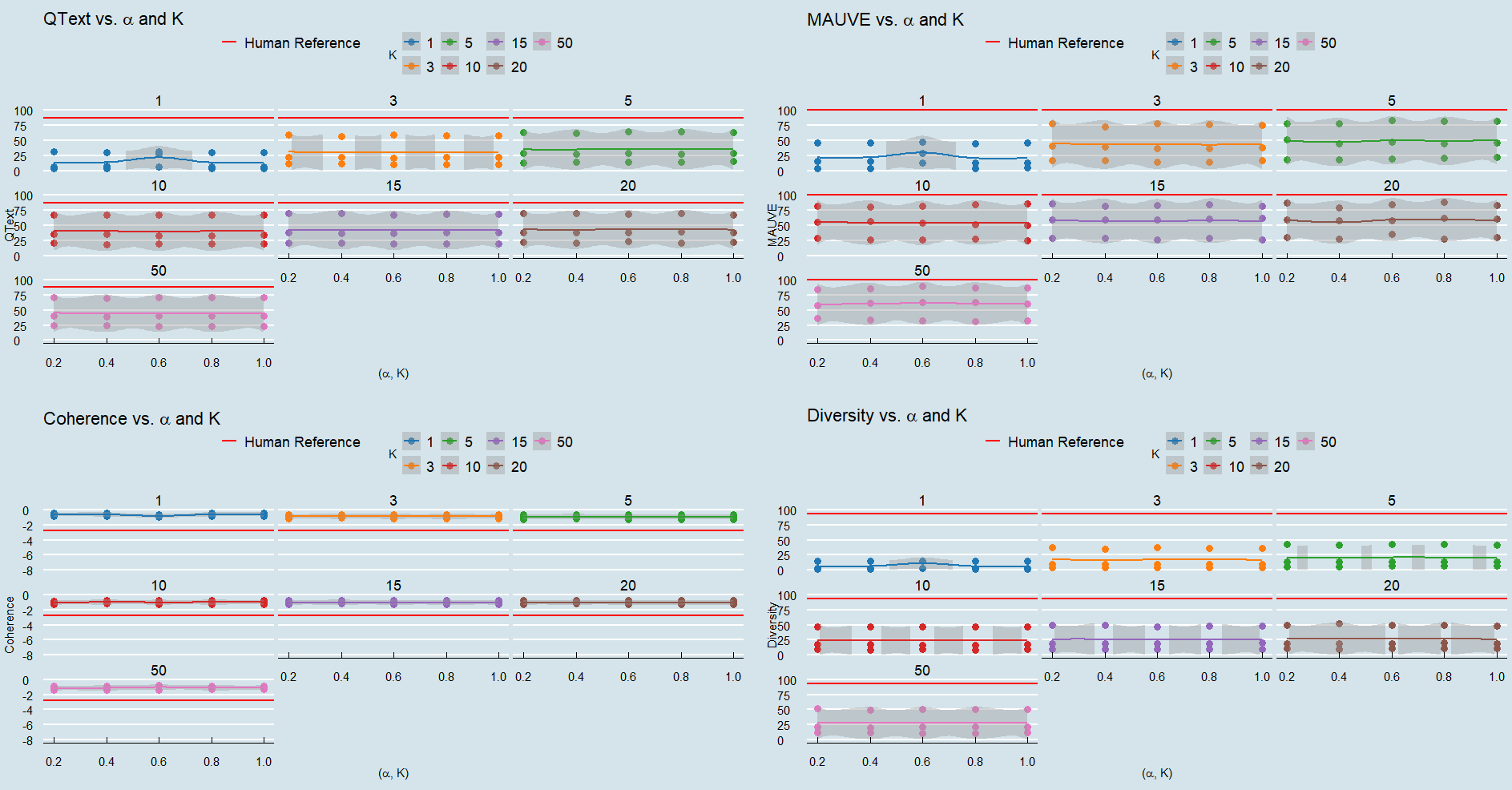}

\caption{Effect of $k$ (fix) and $\alpha$ on metric behavior (model Llama3 visualized).}
\label{fig:csb_metrics_llama}
\end{figure}

\begin{figure}[H]
\centering
\includegraphics[width=1.0\textwidth]
{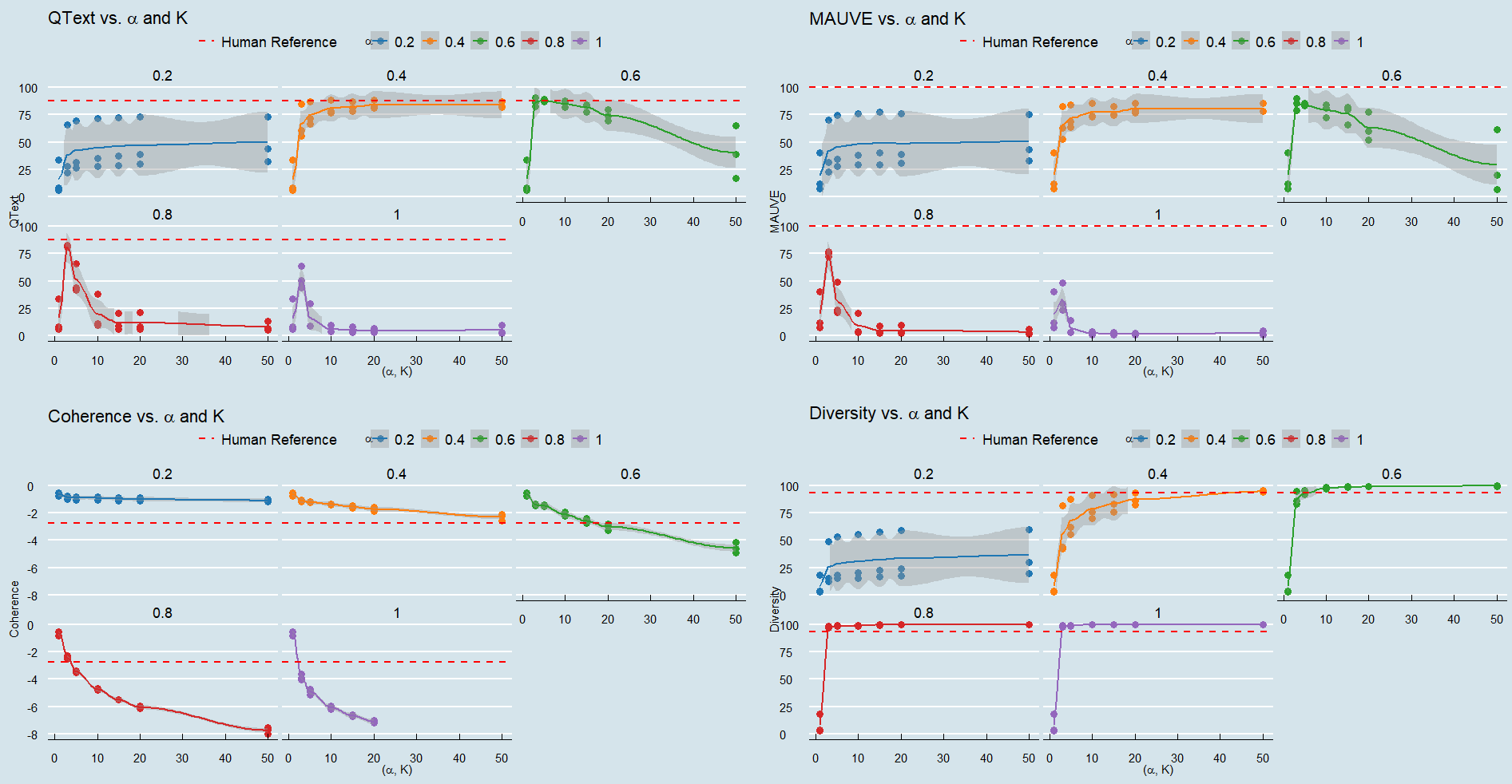}

\caption{Effect of $\alpha$ (fix) and $k$ on metric behavior (model Mistralv03 visualized).}
\label{fig:cs_metrics_mistral}
\end{figure}

\begin{figure}[H]
\centering
\includegraphics[width=1.0\textwidth]
{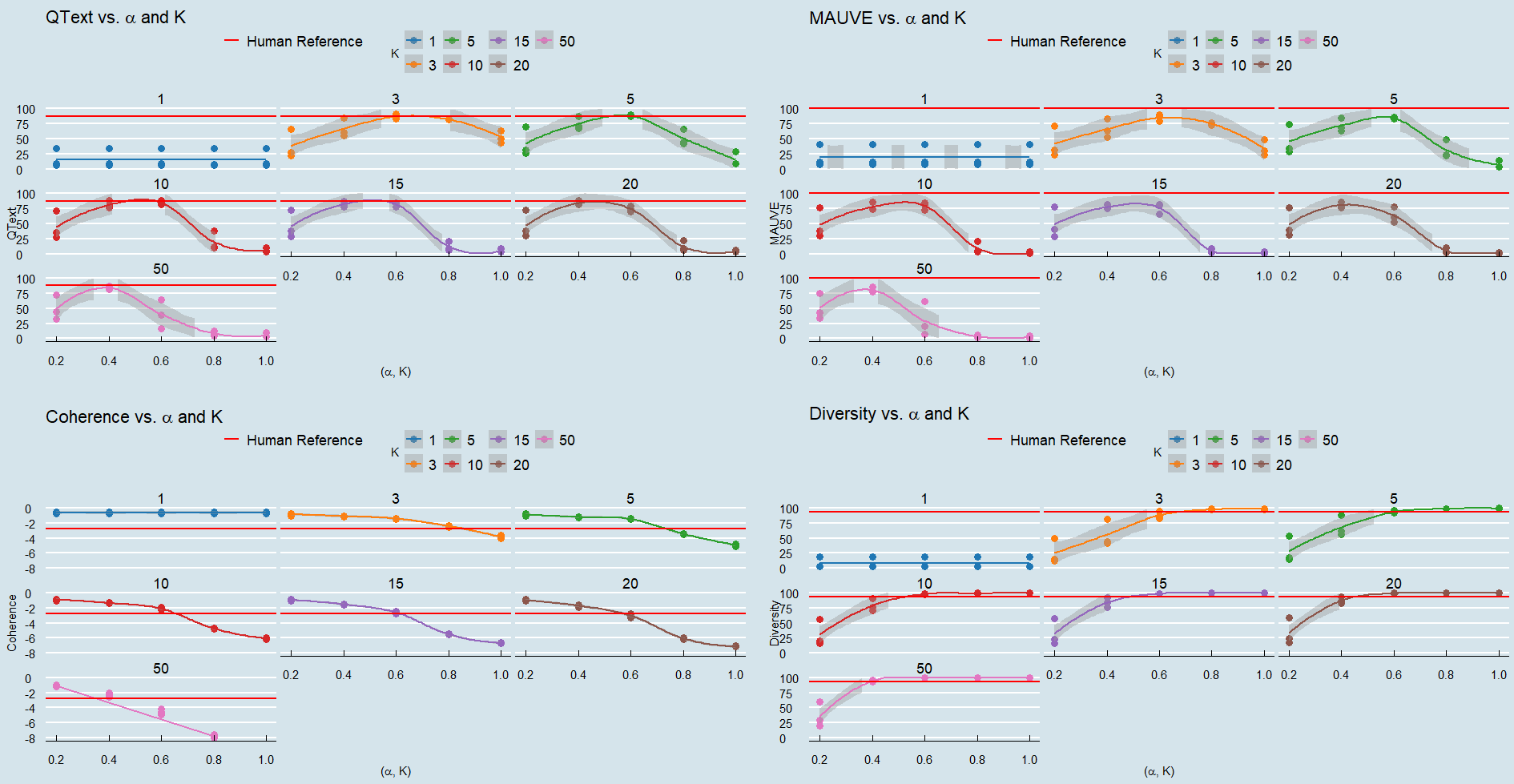}

\caption{Effect of $k$ (fix) and $\alpha$ on metric behavior (model Mistralv03 visualized).}
\label{fig:csb_metrics_mistral}
\end{figure}

\begin{figure}[H]
\centering
\includegraphics[width=1.0\textwidth]
{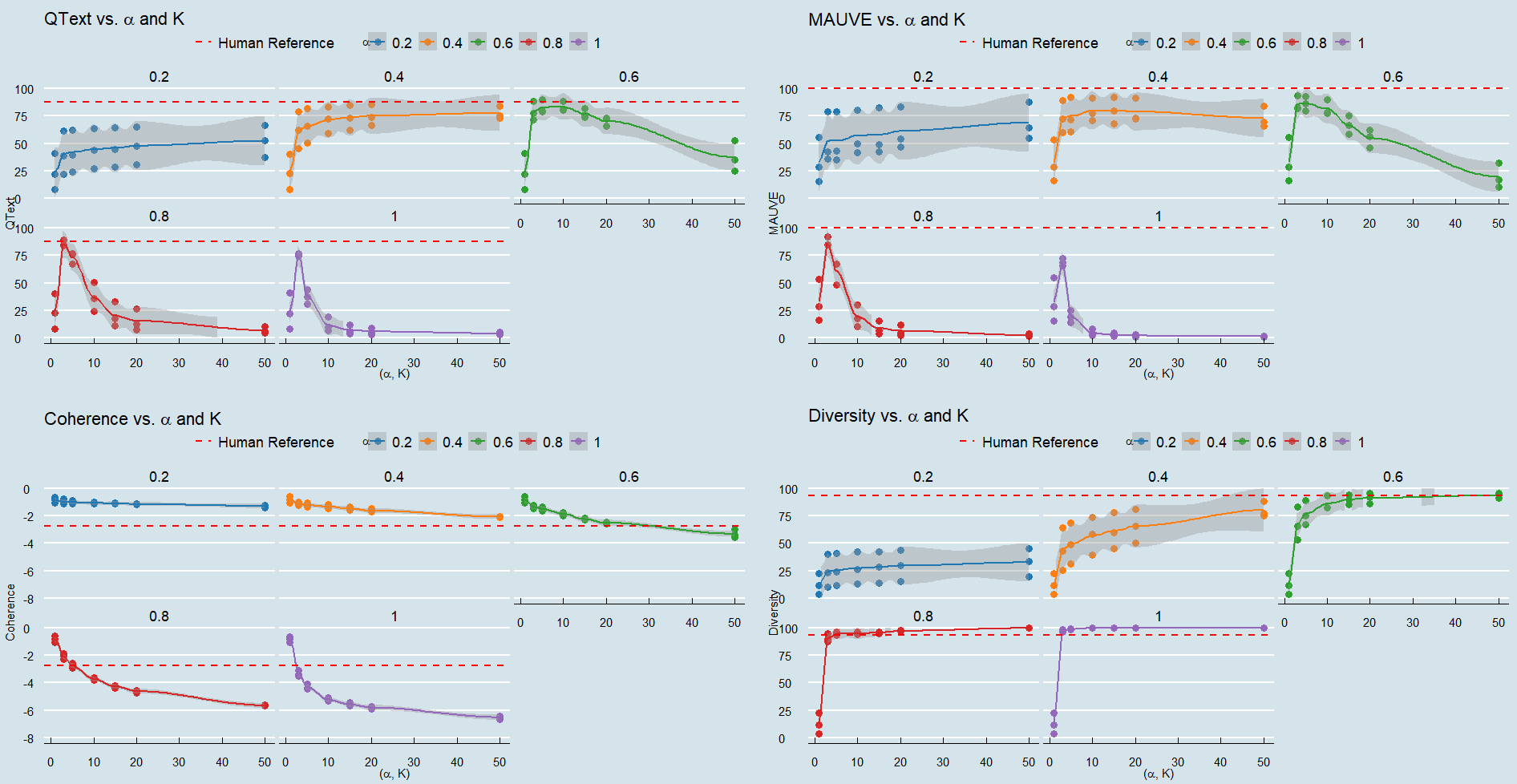}

\caption{Effect of $\alpha$ (fix) and $k$ on metric behavior (model Qwen2 visualized).}
\label{fig:cs_metrics_qwen}
\end{figure}

\begin{figure}[H]
\centering
\includegraphics[width=1.0\textwidth]
{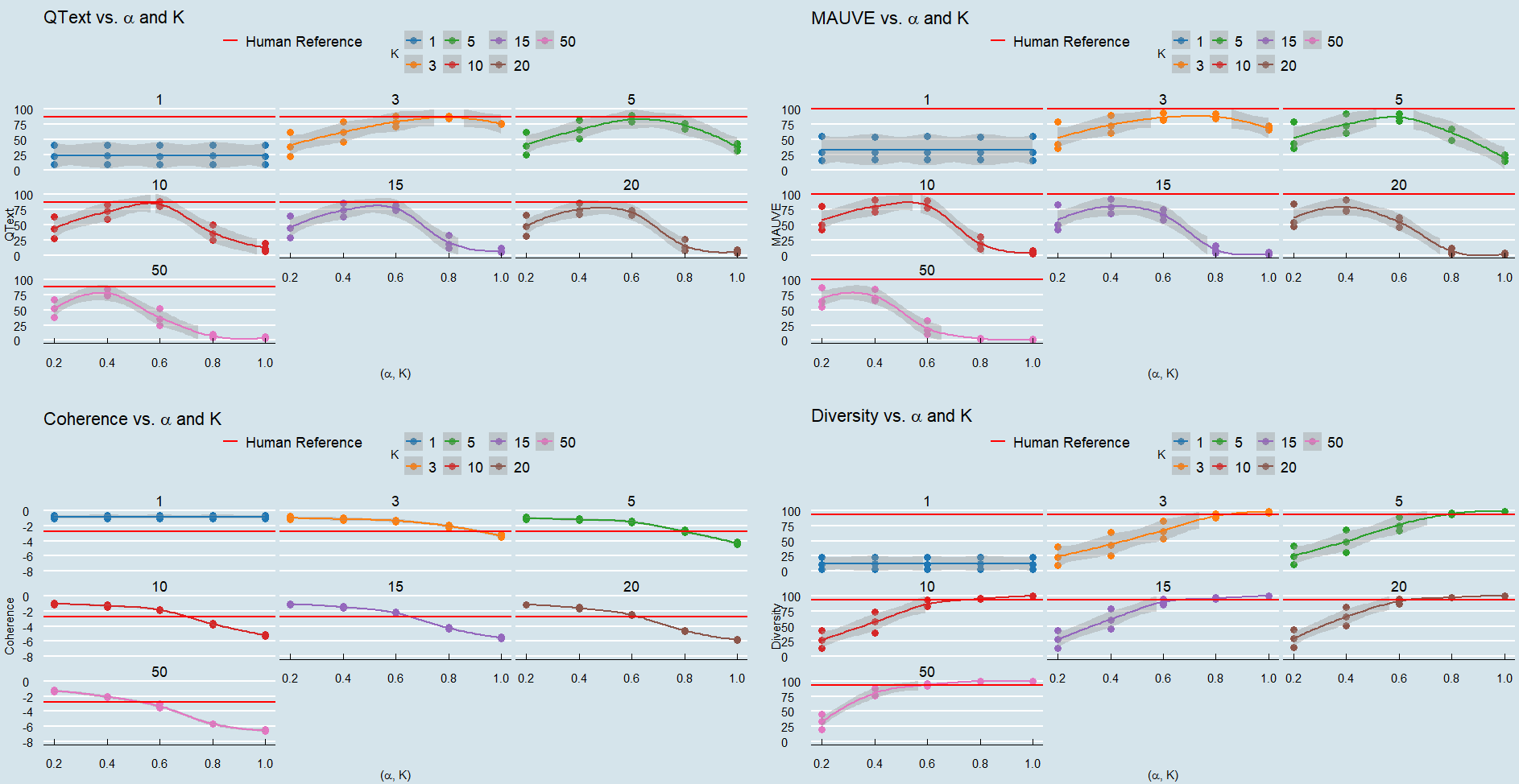}

\caption{Effect of $k$ (fix) and $\alpha$ on metric behavior (model Qwen2 visualized).}
\label{fig:csb_metrics_qwen}
\end{figure}

\begin{figure}[H]
\centering
\includegraphics[width=1.0\textwidth]
{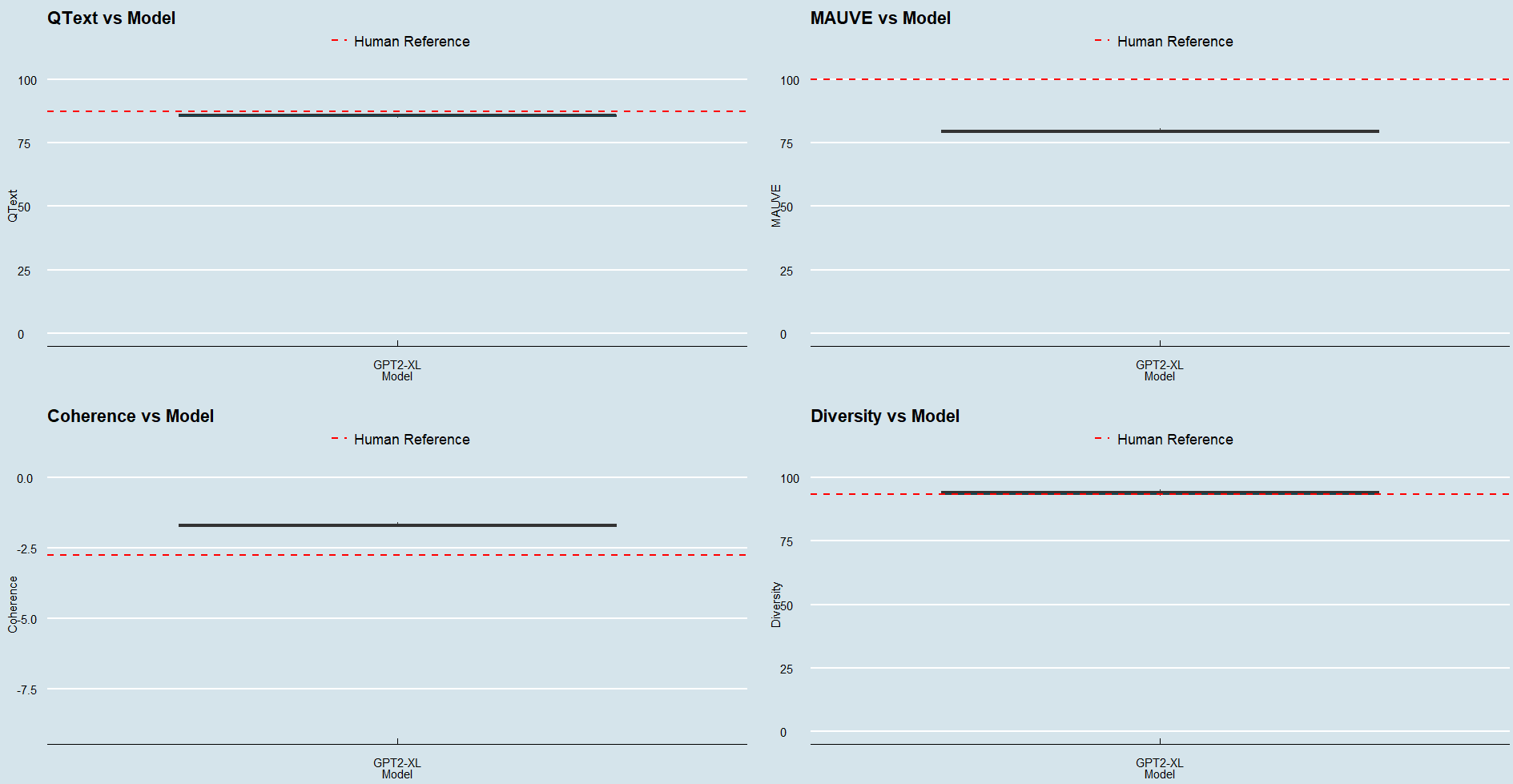}

\caption{Distribution of metric values per model, by using an \textit{Adaptive Contrastive Search} decoding strategy, here we report results for GPT2-XL only.}
\label{fig:acs_metrics_model}
\end{figure}

\begin{figure}[H]
\centering
\includegraphics[width=1.0\textwidth]
{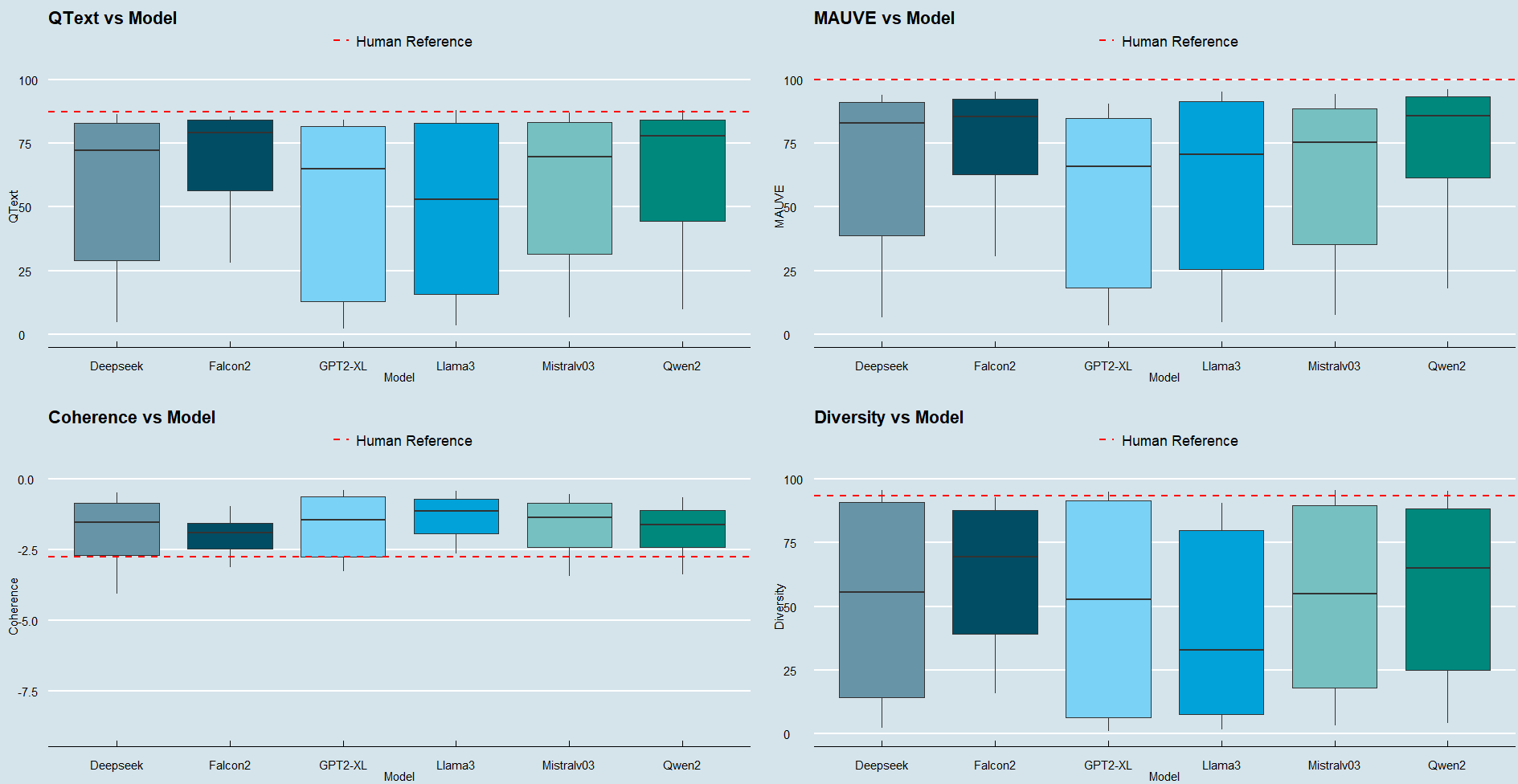}

\caption{Distribution of metric values per model, by using a \textit{Sampling with temperature} decoding strategy.}
\label{fig:swt_boxplot_model}
\end{figure}

\begin{figure}[H]
\centering
\includegraphics[width=1.0\textwidth]
{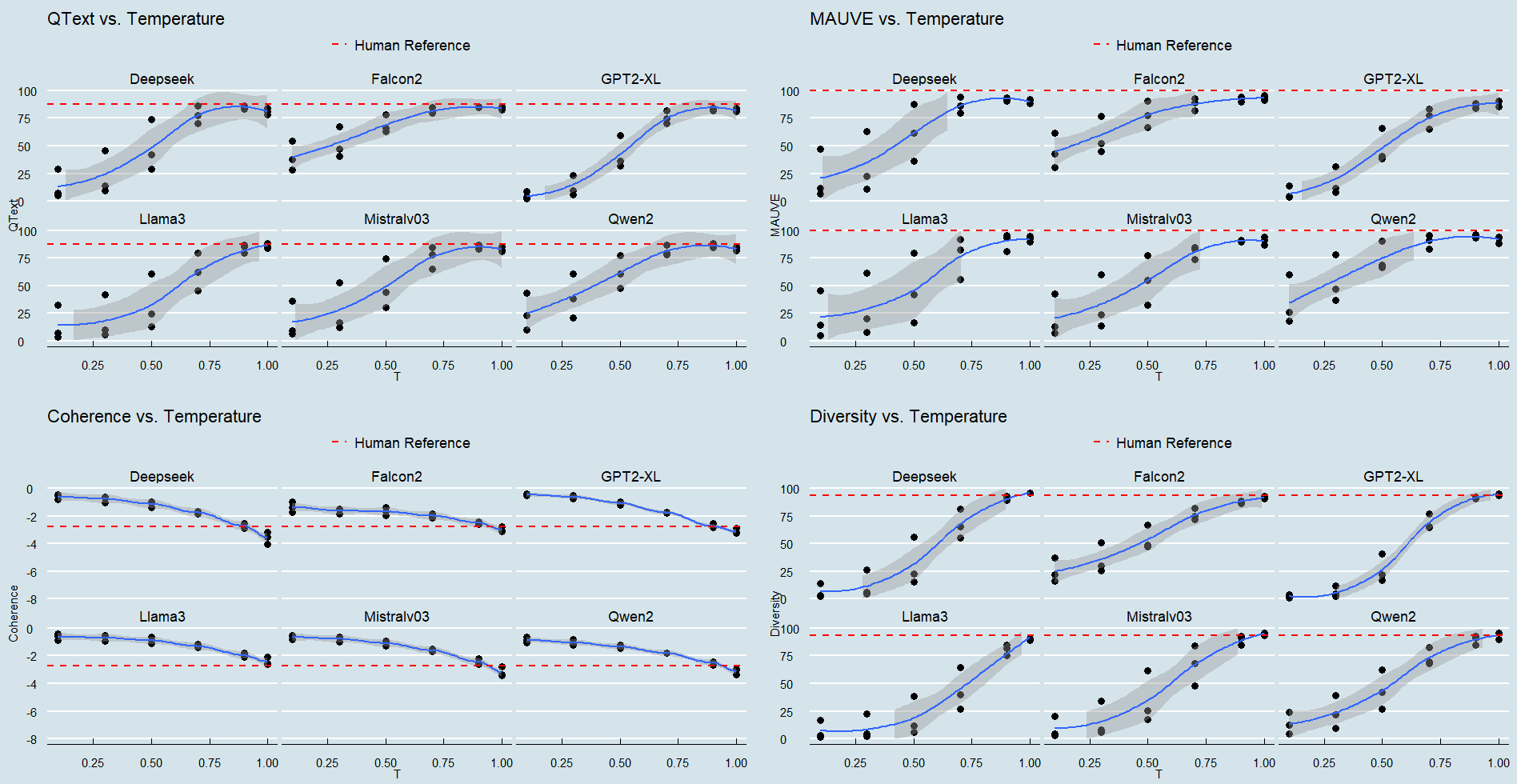}

\caption{Effect of \textit{temperature} on metric behavior, visualized by model.}
\label{fig:swt_metrics_model}
\end{figure}

\begin{figure}[H]
\centering
\includegraphics[width=1.0\textwidth]
{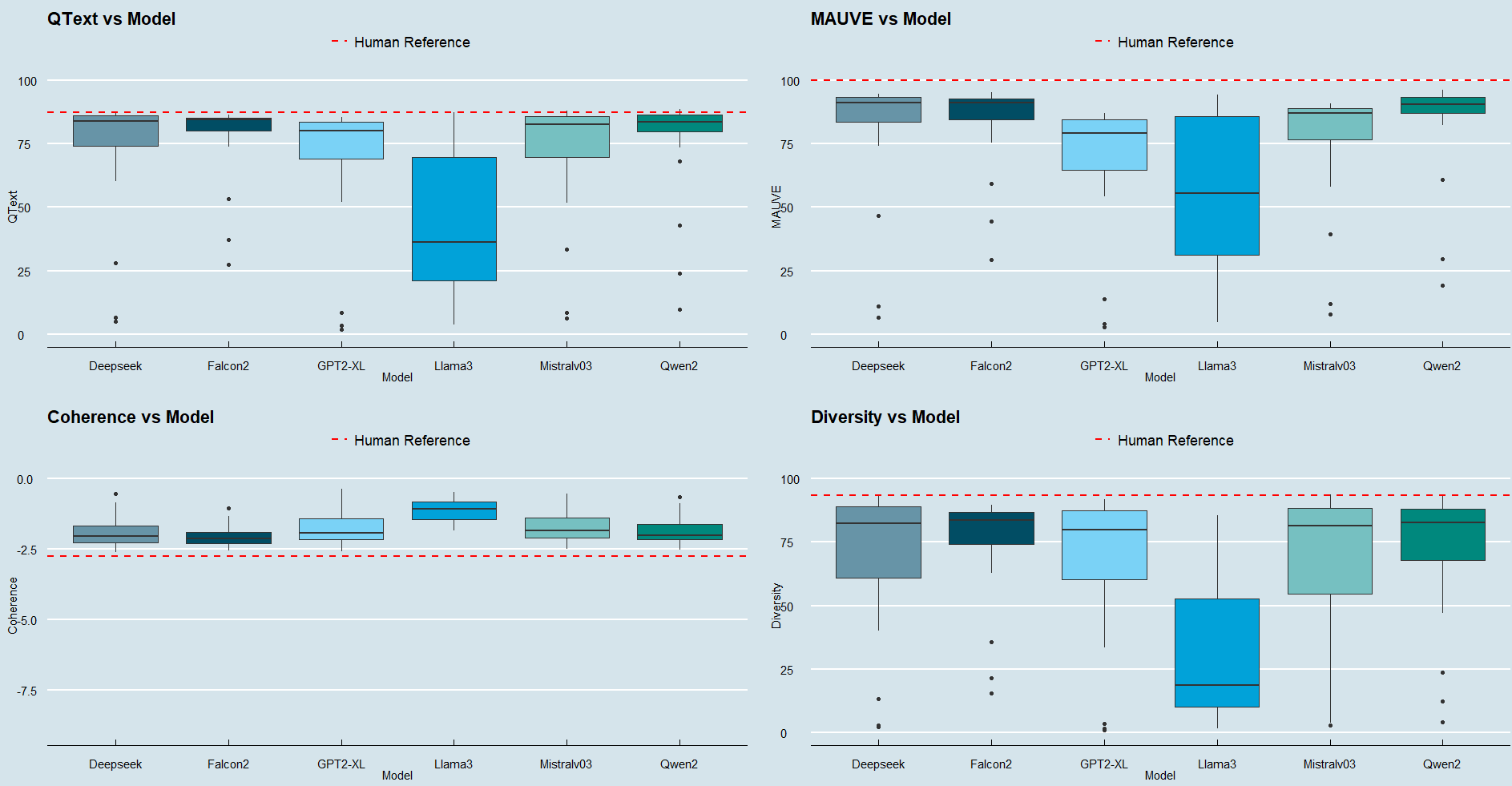}

\caption{Distribution of metric values per model, by using a \textit{Top-$k$ Sampling} decoding strategy.}
\label{fig:topk_boxplot_model}
\end{figure}

\begin{figure}[H]
\centering
\includegraphics[width=1.0\textwidth]
{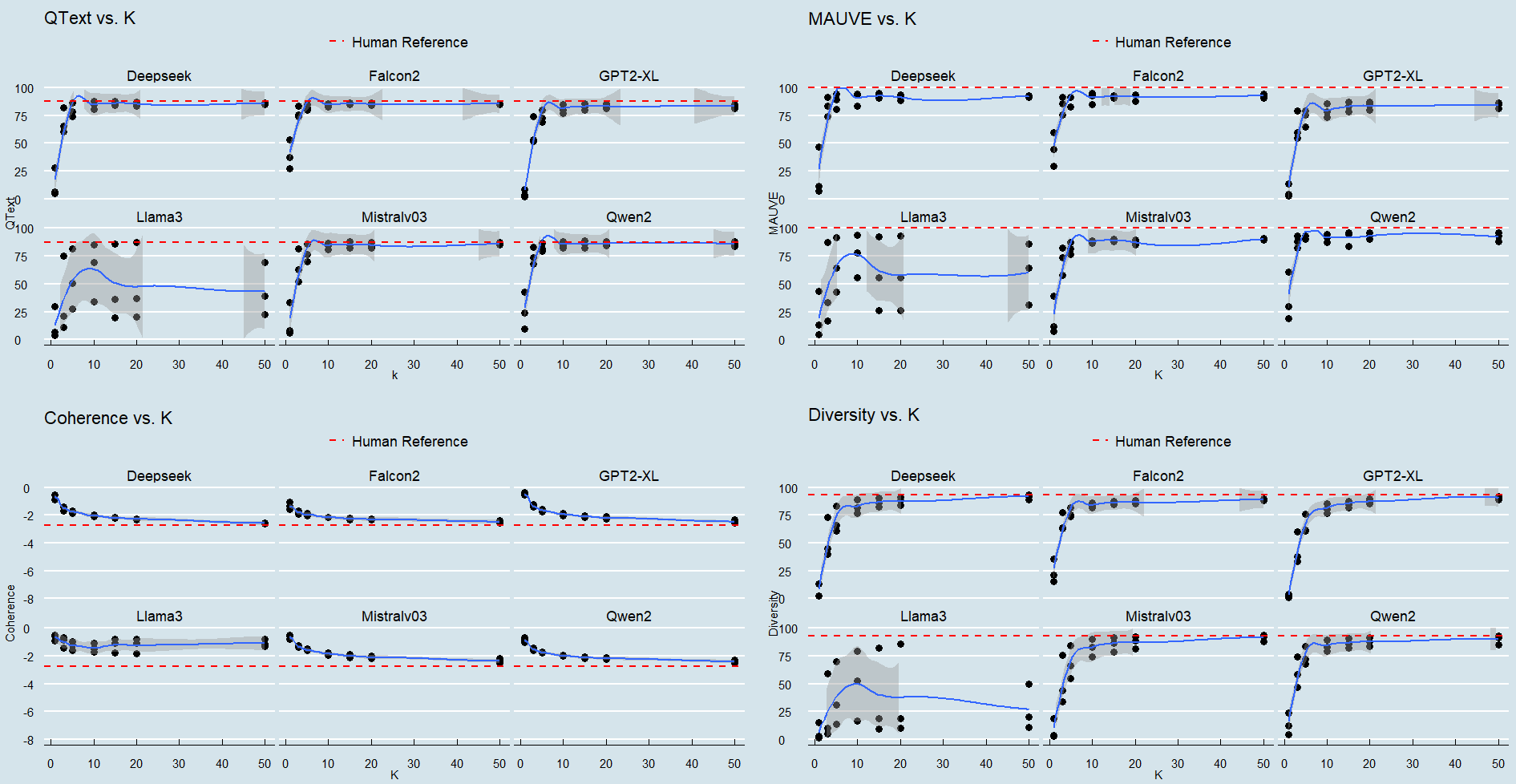}

\caption{Effect of \textit{k} on metric behavior, visualized by model.}
\label{fig:topk_metrics_model}
\end{figure}

\begin{figure}[H]
\centering
\includegraphics[width=1.0\textwidth]
{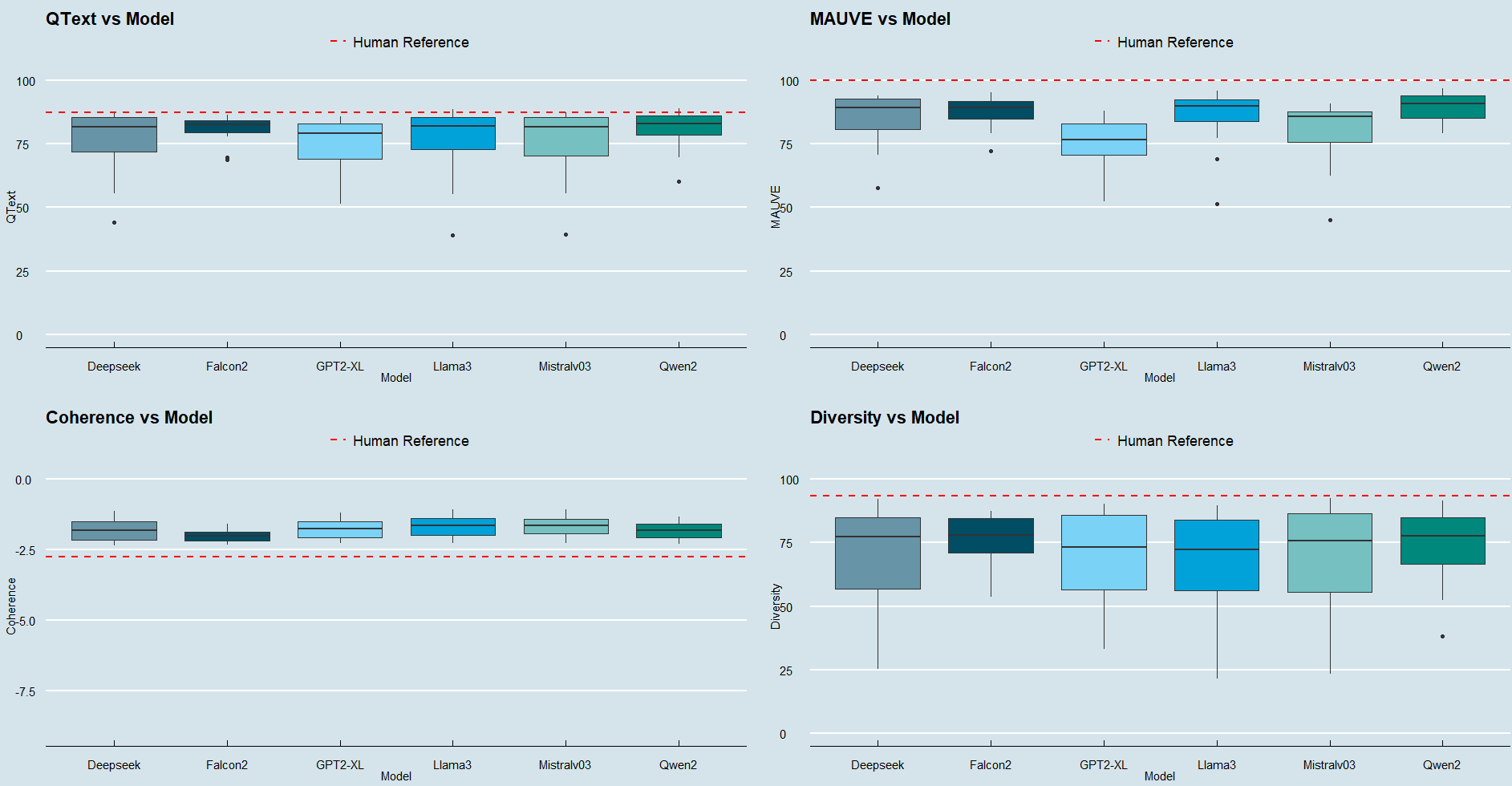}

\caption{Distribution of metric values per model, by using a \textit{Top-$p$ Nucleus Sampling} decoding strategy.}
\label{fig:topp_boxplot_model}
\end{figure}

\begin{figure}[H]
\centering
\includegraphics[width=1.0\textwidth]
{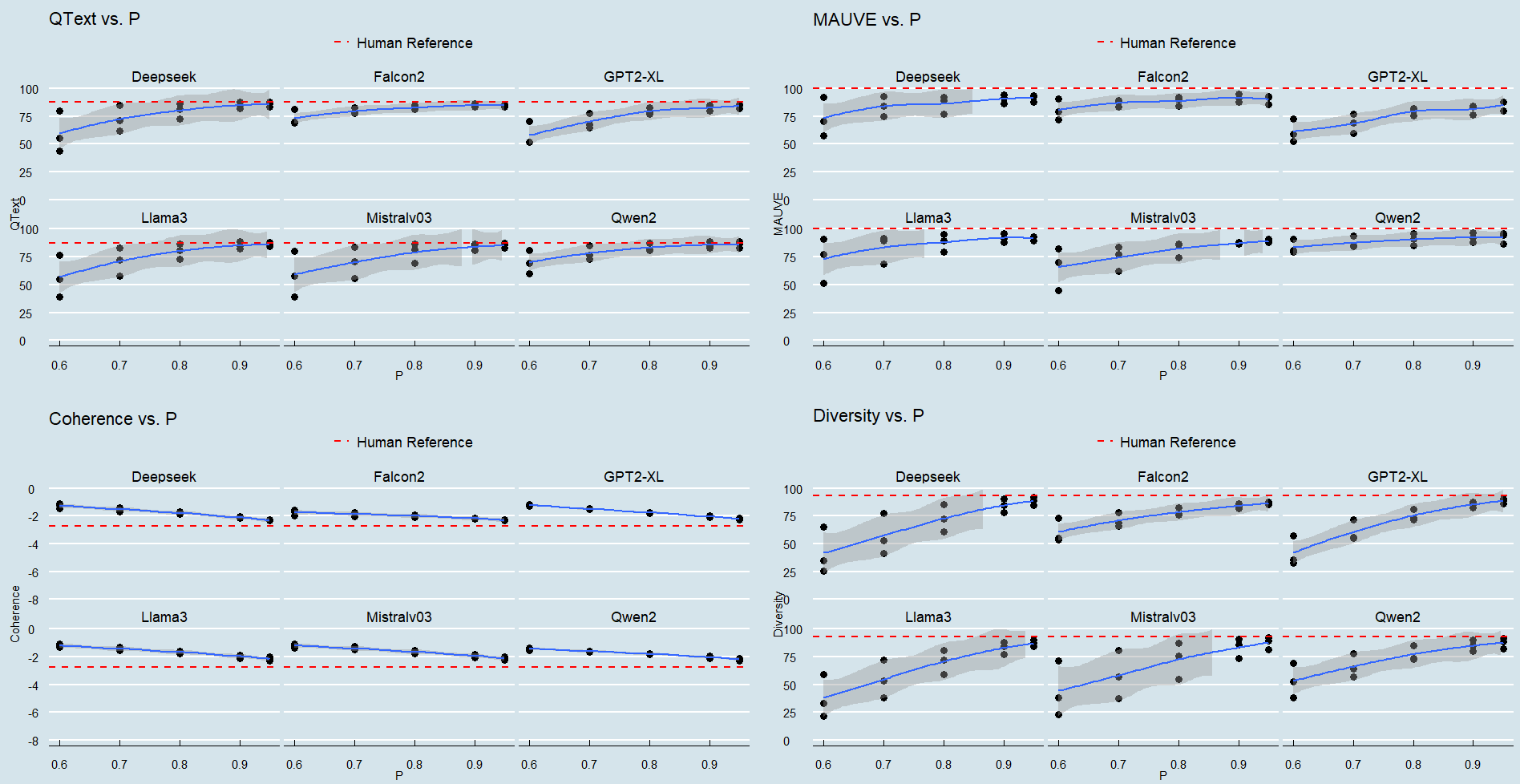}

\caption{Effect of \textit{p} on metric behavior, visualized by model.}
\label{fig:topp_metrics_model}
\end{figure}

\subsection{Correlation of Human evaluation and Automatic metrics}
\label{a:humeval}

\begin{figure}[H]
\centering
\includegraphics[width=\textwidth]
{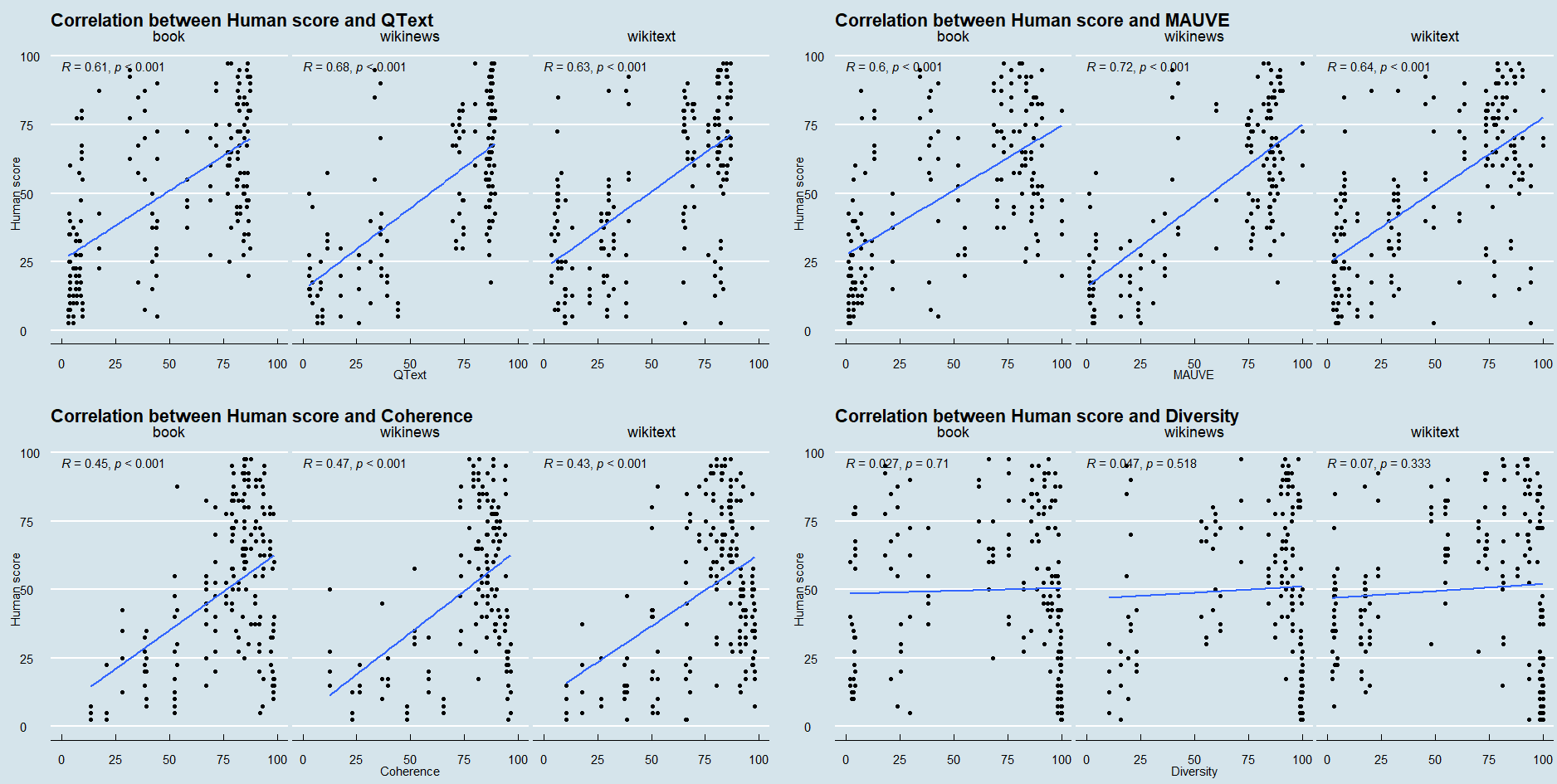}

\caption{Correlation between human judgments and automatic metrics. Human scores show moderate, statistically significant positive correlations with QText and MAUVE. Coherence shows a low, statistically significant positive correlation, while Diversity exhibits a very weak correlation that is not statistically significant.}
\label{fig:correlations_auto_human}
\end{figure}

\clearpage

\onecolumn

\subsection{Human Evaluation Form}
\label{a:humeval}

\begin{figure}[ht]
    \centering
    \adjustbox{rotate=90, max width= 1.1\textheight, max totalheight= 1.1\textwidth}{%
        \includegraphics{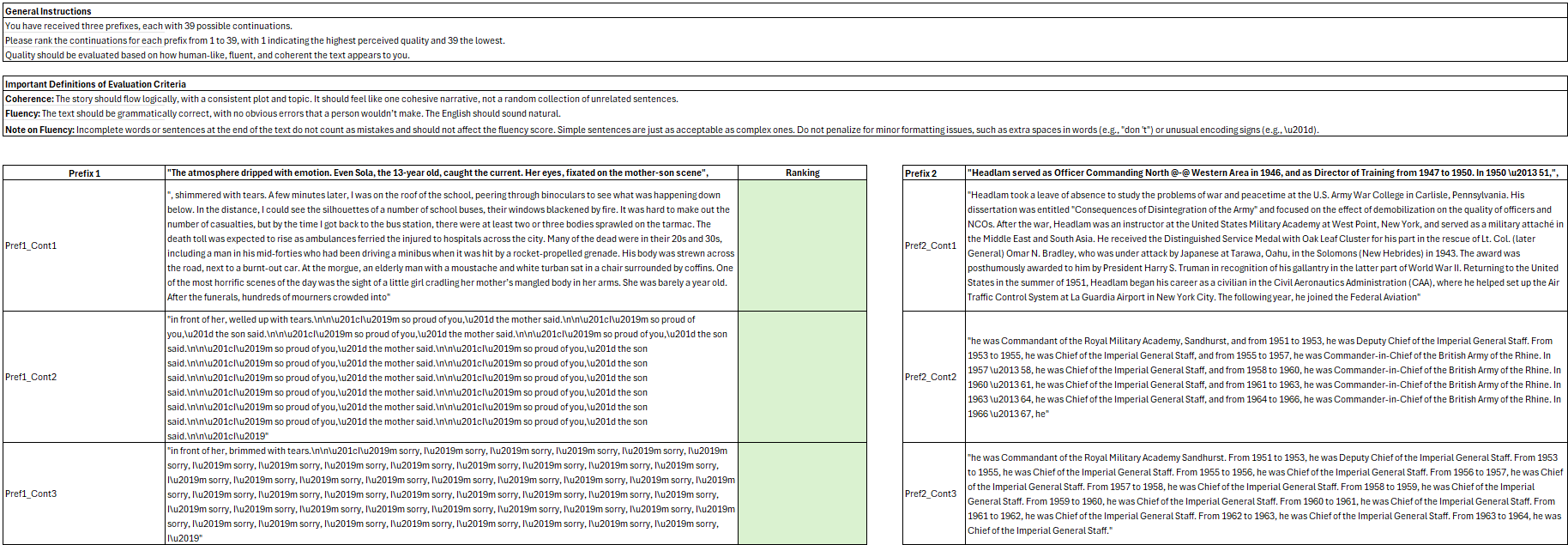}
    }
    \caption{Human evaluation form, including general instructions and definitions for the evaluation criteria.}
    \label{fig:human_evaluation_form}
\end{figure}



\end{document}